\documentclass[twoside]{article}

\usepackage[accepted]{aistats2021}

\usepackage[]{algorithm2e}
\usepackage{amsfonts}
\usepackage{amsmath}
\usepackage{amssymb}
\usepackage{amsthm}
\usepackage{cleveref}
\usepackage{color}
\usepackage{comment}
\usepackage[utf8]{inputenc}
\usepackage{mathtools}
\usepackage{natbib}
\usepackage{url}
\usepackage{cancel}
\allowdisplaybreaks
\usepackage{float}
\usepackage{subcaption}
\RestyleAlgo{ruled}
\usepackage{balance}
\usepackage{lipsum}

\usepackage{etoolbox}
\makeatletter
\patchcmd{\@algocf@start}
  {-1.5em}
  {0pt}
  {}{}
\makeatother

\runningtitle{Optimal Quantisation of Probability Measures Using Maximum Mean Discrepancy} \runningauthor{Teymur, Gorham, Riabiz, Oates}

\newtheorem{theorem}{Theorem}
\newtheorem{remark}{Remark}

\newtheorem{standing}{Standing Assumption}
\newtheorem{lemma}{Lemma}

\def\d{\,\mathrm{d}}

\begin{document}

\twocolumn[

\aistatstitle{Optimal Quantisation of Probability Measures \\Using Maximum Mean Discrepancy}

\aistatsauthor{ Onur Teymur \And Jackson Gorham \And Marina Riabiz \And Chris. J. Oates }

\aistatsaddress{ Newcastle University \\ Alan Turing Institute\And Whisper.ai, Inc. \And King's College London \\ Alan Turing Institute \And Newcastle University \\ Alan Turing Institute } ]

\begin{abstract}
    Several researchers have proposed minimisation of maximum mean
    discrepancy (MMD) as a method to quantise probability measures, i.e., to
    approximate a distribution by a representative point set.
    We consider sequential algorithms that greedily minimise MMD over a discrete candidate set.
    We propose a novel non-myopic algorithm and, in order to both improve statistical efficiency and
    reduce computational cost, we investigate a variant that applies this
    technique to a mini-batch of the candidate set at each iteration.
    When the candidate points are sampled from the target, the consistency
    of these new algorithms---and their mini-batch variants---is established.
    We demonstrate the algorithms on a range of important
    computational problems, including optimisation of nodes in Bayesian cubature and the thinning of Markov chain output.

\end{abstract}

\section{Introduction} \label{sec: intro}
This paper considers the approximation of a probability distribution $\mu$, defined on a set $\mathcal{X}$, by a discrete distribution $\nu = \frac{1}{n} \sum_{i=1}^n \delta(x_i)$, for some representative points $x_i$, where $\delta(x)$ denotes a point mass located at $x \in \mathcal{X}$.
This \emph{quantisation} task arises in many areas including numerical cubature \citep{karvonen2019kernel}, experimental design \citep{chaloner1995bayesian} and Bayesian computation \citep{riabiz2020optimal}. 
To solve the quantisation task one first identifies an optimality criterion, typically a notion of \emph{discrepancy} between $\mu$ and $\nu$, and then develops an algorithm to approximately minimise it.
Classical optimal quantisation picks the $x_i$ to minimise a Wasserstein distance between $\nu$ and $\mu$, which leads to an elegant connection with Voronoi partitions whose centres are the $x_i$ \citep{graf2007foundations}.
Several other discrepancies exist but are less well-studied for the quantisation task.
In this paper we study quantisation with \emph{maximum mean discrepancy}
(MMD), as well as a specific version called \emph{kernel Stein
  discrepancy} (KSD), each of which admit a closed-form expression for a wide class of target distributions $\mu$
\citep[e.g.][]{rustamov2019closed}.

Despite several interesting results, optimal quantisation with MMD remains largely unsolved.
Quasi Monte Carlo (QMC) provides representative point sets that asymptotically minimise MMD \citep{hickernell1998generalized,dick2010digital}; however, these results are typically limited to specific instances of $\mu$ and MMD.\footnote{In \Cref{subsec: MMD} we explain how MMD is parametrised by a \emph{kernel}; the QMC literature typically focuses on $\mu$ uniform on $[0,1]^d$, and $d$-dim tensor products of kernels over $[0,1]$.}
The use of greedy sequential algorithms, in which the point $x_n$ is selected conditional on the points $x_1,\dots,x_{n-1}$ already chosen, has received some attention in the context of MMD---see the recent surveys in \cite{oettershagen2017construction} and \cite{pronzato2018bayesian}.
Greedy sequential algorithms have also been proposed for KSD \citep{chen2018stein,chen2019stein}, as well as a non-greedy sequential algorithm for minimising MMD, called \emph{kernel herding} \citep{chen2012super}.

In certain situations,\footnote{Specifically, the algorithms coincide when the kernel on which they are based is translation-invariant.} the greedy and herding algorithms produce the same sequence of points, with the latter theoretically understood due to its interpretation as a Frank--Wolfe algorithm \citep{bach2012equivalence,lacoste2015sequential}.
Outside the translation-invariant context, empirical studies have shown that greedy algorithms tend to outperform kernel herding \citep{chen2018stein}.
Information-theoretic lower bounds on MMD have been derived in the literature on information-based complexity \citep{novak2008tractability} and in \cite{mak2018support}, who studied representative points that minimise an \emph{energy distance}; the relationship between energy distances and MMD is clarified in \cite{sejdinovic2013equivalence}.

The aforementioned sequential algorithms require that, to select the next point $x_n$, one has to search over the whole set $\mathcal{X}$.
This is often impractical, since $\mathcal{X}$ will typically be an infinite set and may not have useful structure (e.g. a vector space) that can be exploited by a numerical optimisation method.

Extended notions of greedy optimisation, where at each step one seeks to add a point that is merely `sufficiently close' to optimal, were studied for KSD in \cite{chen2018stein}.
\cite{chen2019stein} proposed a stochastic optimisation approach for this purpose. 
However, the global non-convex optimisation problem that must be solved to find the next point $x_n$ becomes increasingly difficult as more points are selected.
This manifests, for example, in the increasing number of iterations required in the approach of \cite{chen2019stein}.

This paper studies sequential minimisation of MMD over a \emph{finite} candidate set, instead of over the whole of $\mathcal{X}$. This obviates the need to use a numerical optimisation routine, requiring only that a suitable candidate set can be produced.
Such an approach was recently described in \cite{riabiz2020optimal}, where an algorithm termed \emph{Stein thinning} was proposed for greedy minimisation of KSD.
Discrete candidate sets in the context of kernel herding were discussed in \cite{chen2012super} and \citet{lacoste2015sequential}, and in \cite{paige16} for the case that the subset is chosen from the support of a discrete target.
\cite{mak2018support} proposed sequential selection from a discrete candidate set to approximately minimise energy distance, but theoretical analysis of this algorithm was not attempted.


The novel contributions of this paper are as follows:
\begin{itemize}

\item We study greedy algorithms for sequential minimisation of MMD, including novel non-myopic algorithms in which multiple points are selected simultaneously. 
These algorithms are also extended to allow for mini-batching of the candidate set. 
Consistency is established and a finite-sample-size error bound is provided.

\item We show how non-myopic algorithms can be cast as integer quadratic programmes (IQP) that can be exactly solved using standard libraries.



\item A detailed empirical assessment is presented, including a study varying the extent of non-myopic selection, up to and including the limiting case in which all points are selected simultaneously. Such non-sequential algorithms require high computational expenditure, and so a semi-definite relaxation of the IQP is considered in the supplement.

\end{itemize}
The remainder of the paper is structured thus. 
In \Cref{sec: background} we provide background on MMD and KSD.
In \Cref{sec: methods} our novel methods for optimal quantisation are presented.
Our empirical assessment, including comparisons with existing methods, is in \Cref{sec: empirical} and our theoretical assessment is in \Cref{sec: theory}.
The paper concludes with a discussion in \Cref{sec: discuss}.
\setlength{\textfloatsep}{10pt}

\section{Background} \label{sec: background}

Let $\mathcal{X}$ be a measurable space and let $\mathcal{P}(\mathcal{X})$ denote the set of probability distributions on $\mathcal{X}$.
First we introduce a notion of discrepancy between two
measures $\mu,\nu \in \mathcal{P}(\mathcal{X})$, and then specialise
this definition to MMD (\Cref{subsec: MMD}) and KSD (\Cref{subsec: KSD}).

For any $\mu,\nu \in \mathcal{P}(\mathcal{X})$ and set $\mathcal{F}$
consisting of real-valued measurable functions on $\mathcal{X}$, we define a
\emph{discrepancy} to be a quantity of the form
\begin{equation} \textstyle
D_\mathcal{F}(\mu,\nu) = \sup_{f \in \mathcal{F}} \left| \int f \d \mu - \int f\d \nu  \right|, \label{eq:IPM}
\end{equation}
assuming $\mathcal{F}$ was chosen so that all integrals in \eqref{eq:IPM} exist.
The set $\mathcal{F}$ is called \emph{measure-determining} if
$D_\mathcal{F}(\mu,\nu) = 0$ implies $\mu=\nu$, and in this case $D_{\mathcal{F}}$ is called an \textit{integral probability metric} \citep{muller1997integral}.
An example is the Wasserstein metric---induced by choosing $\mathcal{F}$ as the set
of $1$-Lipschitz functions defined on $\mathcal{X}$---that is used in classical
quantisation \cite[Thm. 11.8.2]{dudley2018real}.
Next we describe how MMD and KSD are induced from specific
choices of $\mathcal{F}$.

\subsection{Maximum Mean Discrepancy} \label{subsec: MMD}

Consider a symmetric and positive-definite function $k:\mathcal{X} \times  \mathcal{X} \rightarrow \mathbb{R}$, which we call a \emph{kernel}. 
A kernel \emph{reproduces} a Hilbert space of functions $\mathcal{H}$ from $\mathcal{X}\to\mathbb{R}$ if (i) for all $x \in \mathcal{X}$ we have $k(\cdot,x) \in \mathcal{H}$, and (ii) for all $x \in \mathcal{X}$ and $f \in \mathcal{H}$ we have $\langle k(\cdot,x),f \rangle_\mathcal{H} = f(x)$, where $\langle \cdot, \cdot \rangle_{\mathcal{H}}$ denotes the inner product in $\mathcal{H}$. 
By the Moore--Aronszajn theorem \citep{aronszajn1950theory}, there is a one-to-one mapping between the kernel $k$ and the \emph{reproducing kernel Hilbert space} (RKHS) $\mathcal{H}$, which we make explicit by writing $\mathcal{H}(k)$.
A prototypical example of a kernel on $\mathcal{X} \subseteq \mathbb{R}^d$ is the squared-exponential kernel $k(x,y;\ell) = \exp(-\frac{1}{2}\ell^{-2}\Vert x-y\Vert^2 )$, where $\|\cdot\|$ in this paper denotes the Euclidean norm and $\ell > 0$ is a positive scaling constant.

Choosing the set $\mathcal{F}$ in \eqref{eq:IPM} to be the unit ball $\mathcal{B}(k) := \{f \in \mathcal{H}(k) : \langle f , f \rangle_{\mathcal{H}(k)} \leq 1\}$ of the RKHS $\mathcal{H}(k)$ enables the supremum in \eqref{eq:IPM} to be written in closed form and defines the MMD \citep{song2008learning}:
\begin{align}
\mathrm{MMD}_{\mu,k}(\nu)^2 & := D_{\mathcal{B}(k)}(\mu,\nu) \nonumber \\
& \hspace{-50pt} = \textstyle \iint k(x,y) \d\nu(x)\d\nu(y) - 2 \iint k(x,y) \d\nu(x) \d \mu(y) \nonumber \\
& \hspace{40pt} \textstyle + \iint k(x,y) \d\mu(x) \d\mu(y)  \label{eq:MMD}
\end{align}
Our notation emphasises $\nu$ as the variable of interest, since in this paper we aim to minimise MMD over possible $\nu$ for a fixed kernel $k$ and a fixed target $\mu$.
Under suitable conditions on $k$ and $\mathcal{X}$ it can be shown that MMD is a metric on $\mathcal{P}(\mathcal{X})$ (in which case the kernel is called \emph{characteristic}); for sufficient conditions see Section 3 of \cite{sriperumbudur2010hilbert}.
Furthermore, under stronger conditions on $k$, MMD metrises the weak topology on $\mathcal{P}(\mathcal{X})$ \citep[Thms. 23, 24]{sriperumbudur2010hilbert}.
This provides theoretical justification for minimisation of MMD: if $\text{MMD}_{\mu,k}(\nu) \rightarrow 0$ then $\nu \Rightarrow \mu$, where $\Rightarrow$ denotes weak convergence in $\mathcal{P}(\mathcal{X})$.

Evaluation of MMD requires that $\mu$ and $\nu$ are either explicit or can be easily approximated (e.g. by sampling), so as to compute the integrals appearing in \eqref{eq:MMD}.
This is the case in many applications and MMD has been widely used \citep{arbel2019maximum,briol2019statistical,cherief2020mmd}.
In cases where $\mu$ is not explicit, such as when it arises as an intractable posterior in a Bayesian context, KSD can be a useful specialisation of MMD that circumvents integration with respect to $\mu$. We describe this next.

\subsection{Kernel Stein Discrepancy} \label{subsec: KSD}

While originally proposed as a means of proving distributional convergence,
Stein's method \citep{stein1972bound} can be used to circumvent the
integration against $\mu$ required in \eqref{eq:MMD} to calculate the MMD.
Suppose we have an operator $\mathcal{A}_\mu$ defined on a set of
functions $\mathcal{G}$ such that $\int \mathcal{A}_\mu g \d \mu
= 0$ holds for all $g \in \mathcal{G}$. Choosing $\mathcal{F} = \mathcal{A}_\mu\mathcal{G} := \{\mathcal{A}_\mu g :
g \in \mathcal{G}\}$ in \eqref{eq:IPM}, we would then have
$D_\mathcal{\mathcal{A}_\mu\mathcal{G}}(\mu,\nu) = \sup_{g \in \mathcal{G}}
\left| \int \mathcal{A}_\mu g \d \nu \right|$, an expression which no longer involves integrals with respect to $\mu$. 
Appropriate choices for $\mathcal{A}_\mu$ and $\mathcal{G}$ were studied in \cite{GorhamMa15}, who termed $D_\mathcal{\mathcal{A}_\mu\mathcal{G}}$ the
\emph{Stein discrepancy}, and these will now be described.

Assume $\mu$ admits a positive and continuously differentiable density $p_\mu$ on $\mathcal{X} = \mathbb{R}^d$;
let $\nabla$ and $\nabla\cdot$ denote the gradient and divergence
operators respectively; and let $k : \mathbb{R}^d \times \mathbb{R}^d
\rightarrow \mathbb{R}$ be a kernel that is continuously differentiable in each argument. Then take
\begin{align*} 
\mathcal{A}_\mu g & := \nabla\cdot g + u_\mu \cdot g, \qquad u_\mu := \nabla \log p_\mu, \\
\mathcal{G} & := \textstyle \{g:\mathbb{R}^d\rightarrow\mathbb{R}^d: \sum_{i=1}^d \langle g_i , g_i \rangle_{\mathcal{H}(k)} \leq 1 \} .
\end{align*}
Note that $\mathcal{G}$ is the unit ball in the $d$-dimensional tensor product of $\mathcal{H}(k)$.
Under mild conditions on $k$ and $\mu$ \cite[Prop. 1]{gorham2017measuring}, it holds that $\int \mathcal{A}_\mu g \d \mu = 0$ for all $g \in \mathcal{G}$.
The set $\mathcal{A}_\mu \mathcal{G}$ can then be shown  \citep{oates2017control} to be the unit ball $\mathcal{B}(k_\mu)$ in a different RKHS $\mathcal{H}({k_\mu})$ with reproducing kernel
\begin{equation}
\begin{aligned}
	k_\mu(x,y) := & \;  \nabla_x\cdot\nabla_y k(x,y) + \nabla_x k(x,y) \cdot u_\mu(y) \\
	& \hspace{-0.8cm} + \nabla_y k(x,y) \cdot u_\mu(x) + k(x,y) u_\mu(x) \cdot u_\mu(y) ,
	\end{aligned} \label{eq: oates unit ball}
\end{equation}
where subscripts are used to denote the argument upon which a differential operator acts.
Since $k_\mu(x,\cdot) \in \mathcal{H}(k_\mu)$, it follows that $\int k_\mu(x,\cdot) \d \mu(x) = 0$ and from \eqref{eq:MMD} we arrive at the \textit{kernel Stein discrepancy} (KSD)
\begin{equation*} \textstyle
	\text{MMD}_{\mu,k_\mu}(\nu)^2 = \iint k_\mu(x,y) \d\nu(x) \d\nu(y) . 
\end{equation*}
Under stronger conditions on $\mu$ and $k$ it can be shown that KSD controls weak convergence to $\mu$ in $\mathcal{P}(\mathbb{R}^d)$, meaning that if $\text{MMD}_{\mu,k_\mu}(\nu) \rightarrow 0$ then $\nu \Rightarrow \mu$ \citep[Thm. 8]{gorham2017measuring}. 
The description of KSD here is limited to $\mathbb{R}^d$, but constructions
also exist for discrete spaces
\citep{yang2018goodness} and more general Riemannian manifolds \citep{barp2018riemannian,xu2020stein,le2020diffusion}. Extensions
that use other operators $\mathcal{A}_\mu$ \citep{Gorham2016, barp2019minimum} \mbox{have also been studied.}

\section{Methods} \label{sec: methods}

In this section we propose novel algorithms for minimisation of MMD over a finite candidate set.
The simplest algorithm is described in \Cref{subsec: sequential}, and from this we generalise to consider both non-myopic selection of representative points and mini-batching in \Cref{subsec: nonmyopic}.
A discussion of non-sequential algorithms, as the limit of non-myopic algorithms where all points are simultaneously selected, is given in \Cref{subsec: global}.

\subsection{A Simple Algorithm for Quantisation} \label{subsec: sequential}

In what follows we assume that we are provided with a finite candidate set $\{x_i\}_{i=1}^n \subset \mathcal{X}$ from which representative points are to be selected.
Ideally, these candidates should be in regions where $\mu$ is supported, but we defer making any assumptions on this set until the theoretical analysis in \Cref{sec: theory}.
The simplest algorithm that we consider greedily minimises MMD over the candidate set; for each $i$, pick
\begin{equation*}
\textstyle \pi(i) \in\! \underset{j \in \{1,\dots,n\}}{\mathrm{argmin}} \text{MMD}_{\mu,k}\left( \frac{1}{i} \sum_{i'=1}^{i-1} \delta(x_{\pi(i')}) + \frac{1}{i} \delta(x_j) \right) ,
\end{equation*}
to obtain, after $m$ steps, an index sequence $\pi \in \{1,\dots,n\}^m$ and associated empirical distribution $\nu = \frac{1}{m} \sum_{i=1}^m \delta(x_{\pi(i)})$.
(The convention $\sum_{i=1}^0 = 0$ is used.)
Explicit formulae are contained in \Cref{alg: myopic mmd}.
The computational complexity of selecting $m$ points in this manner is $O(m^2 n)$, provided that the integrals appearing in \Cref{alg: myopic mmd} can be evaluated in $O(1)$.
Note that candidate points can be selected more than once.

Theorems \ref{thm: nonmyopic fixed}, \ref{thm: sampled 1} and \ref{thm: sampled 2} in \Cref{sec: theory} provide novel finite-sample-size error bounds for \Cref{alg: myopic mmd} (as a special case of \Cref{alg: non-myopic mmd}). The two main shortcomings of \Cref{alg: myopic mmd} are that (i) the myopic nature of the optimisation may be \emph{statistically inefficient}, and (ii) the requirement to scan through a large candidate set during each iteration may lead to unacceptable \emph{computational cost}.
In \Cref{subsec: nonmyopic} we propose non-myopic and mini-batch extensions to  address these issues.

\begin{algorithm}[t!] 
 \KwData{A set $\{x_i\}_{i=1}^n$, a distribution $\mu$, a kernel $k$ and a number $m \in \mathbb{N}$ of output points}
 \KwResult{An index sequence $\pi \in \{1,\dots,n\}^m$}
 \For{$i=1,\dots,m$}{
 $
\pi(i) \in \underset{j \in \{1,\dots,n\}}{\mathrm{argmin}} \Big[ \frac{1}{2} k(x_j,x_j)+ \sum_{i'=1}^{i-1}k(x_{\pi(i')},x_j) 
$

\vspace{-5pt}
\hspace{133pt} $ \textstyle - i \!\int k(x,x_j) \mathrm{d}\mu(x) \Big] 
$
\vspace{-15pt}
 }
 \caption{Myopic minimisation of MMD}
 \label{alg: myopic mmd}
\end{algorithm}

\subsection{Generalised Sequential Algorithms} \label{subsec: nonmyopic}

In \Cref{subsubsec: IP} we describe a non-myopic extension of \Cref{alg: myopic mmd}, where multiple points are simultaneously selected at each step.
The use of non-myopic optimisation is impractical when a large candidate set is used, and therefore we explain how mini-batches from the candidate set can be employed in \Cref{subsec: minibatch}.

\subsubsection{Non-Myopic Minimisation} \label{subsubsec: IP}

Now we consider the simultaneous selection of $s > 1$ representative points from the candidate set at  each step. This leads to the non-myopic algorithm
$$
\hspace{-5pt} \textstyle \pi(i,\cdot)\! \in\!\!\! \underset{S \in \{1,\dots,n\}^s}{\mathrm{argmin}}\! \text{MMD}_{\mu,k}\Big( \frac{1}{is} \sum_{i'=1}^{i-1} \sum_{j=1}^s \delta(x_{\pi(i',j)}) 
$$

\vspace{-30pt} 
$$ \textstyle
\hspace{160pt} + \frac{1}{is} \sum_{j \in S} \delta(x_j) \Big) ,
$$
whose output is a bivariate index $\pi \in \{1,\dots,n\}^{m \times s}$, together with the associated empirical distribution $\nu = \frac{1}{ms} \sum_{i=1}^m \sum_{j=1}^s \delta(x_{\pi(i,j)})$.
Explicit formulae are contained in \Cref{alg: non-myopic mmd}.
The computational complexity of selecting $ms$ points in this manner is $O(m^2s^2n^s)$, which is larger than \Cref{alg: myopic mmd} when $s > 1$.
Theorems \ref{thm: nonmyopic fixed}, \ref{thm: sampled 1} and \ref{thm: sampled 2} in \Cref{sec: theory} provide novel finite-sample-size error bounds for \Cref{alg: non-myopic mmd}.

Despite its daunting computational complexity, we have found that it is practical to exactly implement \Cref{alg: non-myopic mmd} for moderate values of $s$ and $n$ by casting each iteration of the algorithm as an instance of a constrained \emph{integer quadratic programme} (IQP) \citep[e.g.][]{wolsey21}, so that state-of-the-art discrete optimisation methods can be employed.
To this end, we represent the indices $S \subset \{1,\dots,n\}^s$ of the $s$ points to be selected at iteration $i$ as a vector $v \in \{0,\dots,s\}^n$ whose $j$th element indicates the number of copies of $x_j$ that are selected, and where $v$ is constrained to satisfy $\sum_{j=1}^n v_j = s$. 
It is then an algebraic exercise to recast an optimal subset $\pi(i,\cdot)$ as the solution to a constrained IQP:

\break
\vspace*{-4em}
\begin{gather}	
\textstyle
\underset{v \in \mathbb{N}_0^s}{\mathrm{argmin}} \ \tfrac{1}{2}v^\top K v + c^{i\top} v \quad \text{s.t.} \quad \mathbf{1}^\top v = s \label{eq: bqp formulation} \\
K_{j,j'} := k(x_j,x_{j'}), \ \ \ \ \mathbf{1}_j := 1\  \text{for}\  j=1,\dots,n, \nonumber \\[0.2em]
c^i_j := \textstyle \sum_{i'=1}^{i-1}\sum_{j'=1}^s k(x_{\pi(i',j')},x_j) - is\int k(x,x_j)\d \mu(x) \nonumber
\end{gather}
\begin{remark} \label{rem: binary}
If one further imposes the constraint $v_i \in \{0,1\}$ for all $i$, so that each candidate  may be selected at most once, then the resulting \emph{binary quadratic programme} (BQP) is equivalent to the \emph{cardinality constrained $k$-partition} problem from discrete optimisation, which is known to be NP-hard \citep{Rendl_2012}. 
(The results we present do \emph{not} impose this constraint.)
\end{remark}

\begin{algorithm}[t!] 
 \KwData{A set $\{x_i\}_{i=1}^n$, a distribution $\mu$, a kernel $k$, a number of points to select per iteration $s \in \mathbb{N}$ and a total number of iterations $m \in \mathbb{N}$}
 \KwResult{An index sequence $\pi \in \{1,\dots,n\}^{m \times s}$}
 \For{$i=1,\dots,m$}{
$
\pi(i,\cdot) \in \underset{S \in \{1,\dots,n\}^s }{\mathrm{argmin}} \Big[ \textstyle \frac{1}{2} \sum_{j,j' \in S} k(x_j,x_{j'}) 
$

\vspace{00pt}
$ \hspace{50pt} \textstyle
+ \sum_{i'=1}^{i-1} \sum_{j=1}^s \sum_{j' \in S} k(x_{\pi(i',j)},x_{j'})  
$

\hspace{100pt} $ \textstyle - i s \sum_{j \in S} \int k(x,x_j) \mathrm{d}\mu(x) \Big] 
$
\vspace{-15pt}
 }
 \caption{Non-myopic minimisation of MMD}
 \label{alg: non-myopic mmd}
\end{algorithm}

\begin{figure*}[t]
	\includegraphics[width=0.24\textwidth]{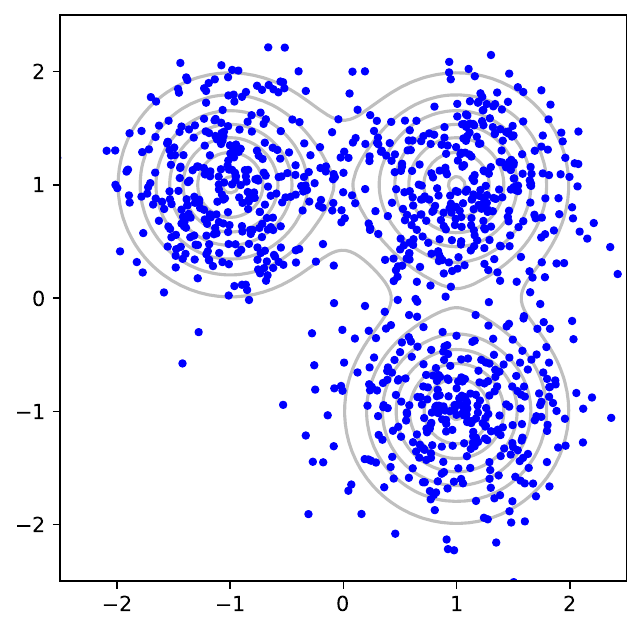}
	\includegraphics[width=0.24\textwidth]{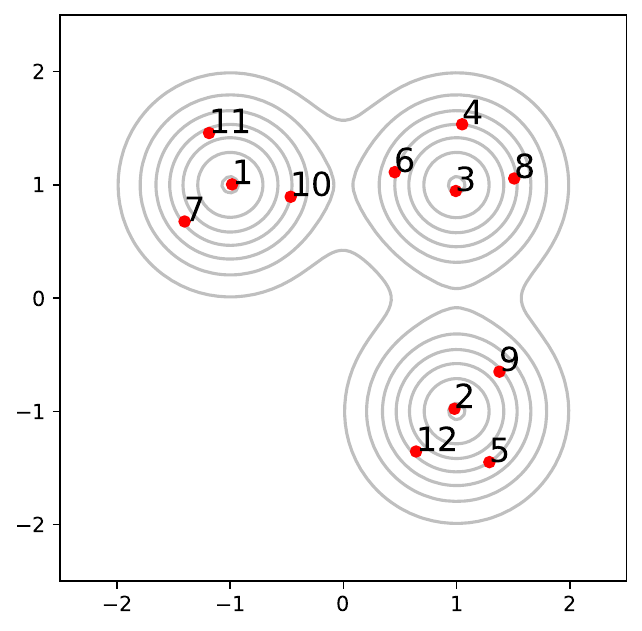}
	\includegraphics[width=0.24\textwidth]{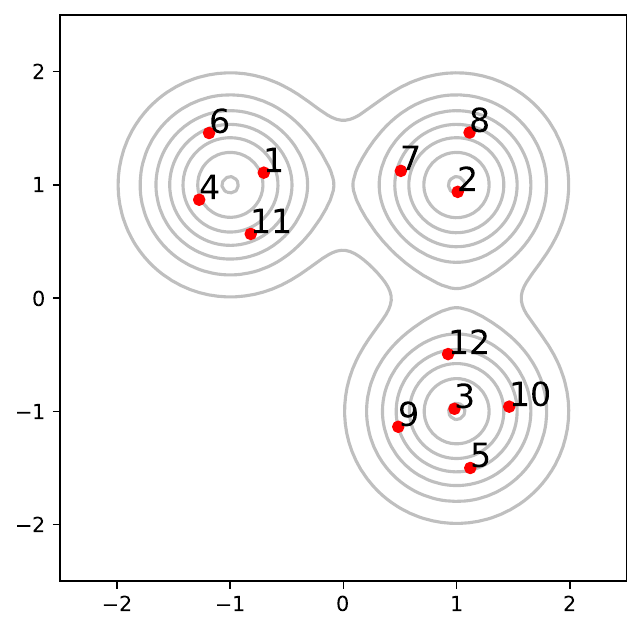}
	\includegraphics[width=0.24\textwidth]{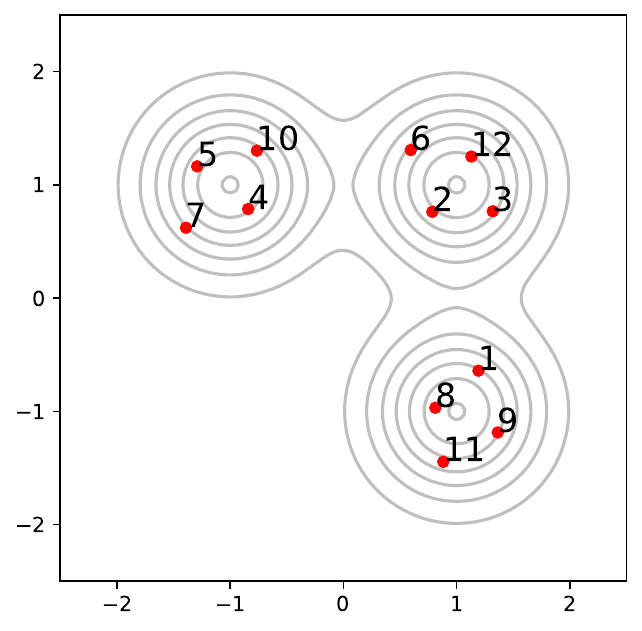}  
	\centering
	\caption{Quantisation of a Gaussian mixture model using MMD. 
	A candidate set of 1000 independent samples (left), from which 12 representative points were selected using: the myopic method (centre-left); non-myopic selection, picking 4 points at a time (centre-right), and by simultaneous selection of all 12 points (right). 
	Simulations were conducted using Algorithms 1 and 2, with a Gaussian kernel whose length-scale was $\ell=0.25$.
	}
 \label{fig: visual example}
\end{figure*}

\subsubsection{Mini-Batching} \label{subsec: minibatch}

The exact solution of \eqref{eq: bqp formulation} is practical only for moderate values of $s$ and $n$. 
This motivates the idea of considering only a subset of the $n$ candidates at each iteration, a procedure we call \emph{mini-batching} and inspired by the similar idea from stochastic optimisation.
There are several ways that mini-batching can be performed, but here we simply state that candidates denoted $\{x_j^i\}_{j=1}^b$ are considered during the $i$th iteration, with the mini-batch size denoted by $b \in \mathbb{N}$.
The non-myopic algorithm for minimisation of MMD with mini-batching is then
\begin{equation*}
\begin{aligned}
\textstyle \pi(i,\cdot)\! &\textstyle \in\!\!\! \underset{S \in \{1,\dots,b\}^s}{\mathrm{argmin}}\! \text{MMD}_{\mu,k}\Big( \frac{1}{is} \sum_{i'=1}^{i-1} \sum_{j =1}^s \delta(x_{\pi(i',j)}^{i'} )\\[-8pt]
& \hspace{130pt} + \textstyle \frac{1}{is} \sum_{j \in S} \delta(x_j^i) \Big)
\end{aligned}
\end{equation*}
Explicit formulae are contained in \Cref{alg: non-myopic mmd minibatch}.
The complexity of selecting $ms$ points in this manner is $O(m^2s^2b^s)$, which is smaller than \Cref{alg: non-myopic mmd} when $b < n$.
As with \Cref{alg: non-myopic mmd}, an exact IQP formulation can be employed.
\Cref{thm: mini batch} provides a novel finite-sample-size error bound for \Cref{alg: non-myopic mmd minibatch}.

\begin{algorithm}[t!] 
 \KwData{A set $\{\{x_j^i\}_{j=1}^b\}_{i=1}^m$ of mini-batches, each of size $b\in \mathbb{N}$, a distribution $\mu$, a kernel $k$, a number of points to select per iteration $s \in \mathbb{N}$, and a  number of iterations $m \in \mathbb{N}$}
 \KwResult{An index sequence $\pi \in \{1,\dots,n\}^{m \times s}$}

 \For{$i=1,\dots,m$}{
$
\pi(i,\cdot) \in \underset{S \in \{1,\dots,b\}^s }{\mathrm{argmin}} \Big[ \textstyle \frac{1}{2} \sum_{j,j' \in S} k(x_j^i,x_{j'}^i) 
$

\vspace{-0pt}
$ \hspace{45pt} \textstyle
+ \sum_{i'=1}^{i-1} \sum_{j=1}^s \sum_{j' \in S} k(x_{\pi(i',j)}^{i'},x_{j'}^i)  
$

\hspace{100pt} $ \textstyle - i s \sum_{j \in S} \int k(x,x_j^i) \mathrm{d}\mu(x) \Big] 
$
\vspace{-15pt}
 }
 
 \caption{Non-myopic minimisation of MMD with mini-batching}
 \label{alg: non-myopic mmd minibatch}
\end{algorithm}

\subsection{Non-Sequential Algorithms} \label{subsec: global}

Finally we consider the limit of the non-myopic Algorithm \ref{alg: non-myopic mmd}, in which all $m$ representative points are simultaneously selected in a single step:

\break
\vspace*{-3em}
\begin{align}
\textstyle \pi \in \underset{S \in \{1,\dots,n\}^m}{\mathrm{argmin}} \text{MMD}_{\mu,k}\Big( \frac{1}{m} \sum_{i \in S} \delta(x_{\pi(i)}) \Big) \label{eq: optimal MMD}
\end{align}
The index set $\pi$ can again be recast as the solution to an IQP and the associated empirical measure $\nu = \frac{1}{m} \sum_{i=1}^m \delta(x_{\pi(i)})$ provides, by definition, a value for $\text{MMD}_{\mu,k}(\nu)$ that is at least as small as any of the methods so far described (thus satisfying the same error bounds derived in Theorems \ref{thm: nonmyopic fixed}--\ref{thm: sampled 2}).

However, it is only practical to exactly solve \eqref{eq: optimal MMD} for small $m$ and thus, to arrive at a practical algorithm, we consider approximation of \eqref{eq: optimal MMD}.
There are at least two natural ways to do this.
Firstly, one could run a numerical solver for the IQP formulation of \eqref{eq: optimal MMD} and terminate after a fixed computational limit is reached; the solver will return a feasible, but not necessarily optimal, solution to the IQP.
An advantage of this approach is that no further methodological work is required.
Alternatively, one could employ a convex relaxation of the intractable IQP, which introduces an approximation error that is hard to quantify but leads to a convex problem that may be exactly soluble at reasonable cost.
We expand on the latter approach in Appendix \ref{appendix: sdr}, with preliminary empirical comparisons.

\section{Empirical Assessment} \label{sec: empirical}

This section presents an empirical assessment\footnotemark\ 
of Algorithms \ref{alg: myopic mmd}--\ref{alg: non-myopic mmd minibatch}.
\footnotetext{Our code is written in \texttt{Python} and is available at \url{https://github.com/oteym/OptQuantMMD}} 
Two regimes are considered, corresponding to high compression (small $sm/n$; \Cref{subsec: BC}) and low compression (large $sm/n$; \Cref{subsec: MCMC}) of the target.
These occur, respectively, in applications to Bayesian cubature and thinning of Markov chain output. In Section \ref{sec: continuous comparison}, we compare our method to a variety of others based on optimisation in continuous spaces, augmenting a study in \cite{chen2019stein}. For details of the kernels used, and a sensitivity analysis for the kernel parameters, see Appendix \ref{sec: extra experiments}.

Figure \ref{fig: visual example} illustrates how a non-myopic algorithm may outperform a myopic one. A candidate set was constructed using $1000$ independent samples from a test measure, and 12 representative points selected using the myopic (Alg. 1 with $m=12$), non-myopic (Alg. 2 with $m=3$ and $s=4$), and non-sequential (Alg. 2 with $m=1$ and $s=12$) approaches. After choosing the first three samples close to the three modes, the myopic method then selects points that temporarily worsen the overall approximation; note in particular the placement of the fourth point. The non-myopic methods do not suffer to the same extent: choosing 4 points together gives better approximations after each of 4, 8 and 12 samples have been chosen ($s=4$ was chosen deliberately so as to be co-prime to the number of mixture components, 3). Choosing all 12 points at once gives an even better approximation.

\begin{figure*}[t!]
\centering
\vspace{-0.5em}
\begin{minipage}[b]{0.31\textwidth}
	\hspace{0.2cm}\includegraphics[width=0.95\textwidth,height=0.86\textwidth]{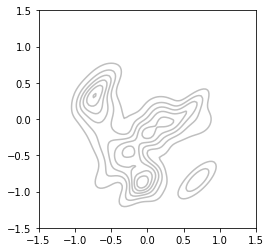}
	
	\hspace{-0.2cm}\includegraphics[width=1.1\textwidth,height=0.95\textwidth]{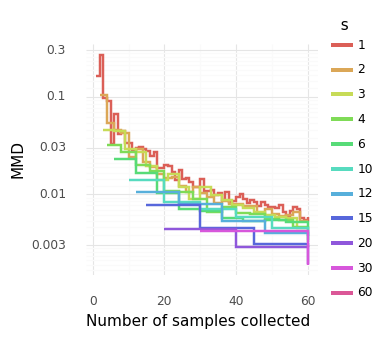}
	\caption{Synthetic data model formed of a mixture of 20 bivariate Gaussians (top). Effect of varying the number $s$ of simultaneously-chosen points on MMD when choosing 60 from 1000 independently sampled points (bottom).}\label{fig: vary batch}
	\end{minipage} \quad \vrule\quad
	\begin{minipage}[b]{0.64\textwidth}
	\includegraphics[width=0.48\textwidth,height=0.44\textwidth]{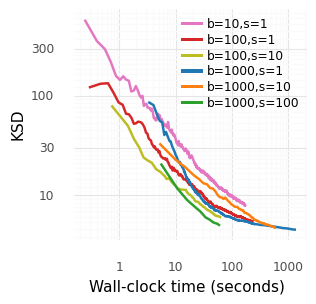}
	\includegraphics[width=0.5\textwidth,height=0.44\textwidth]{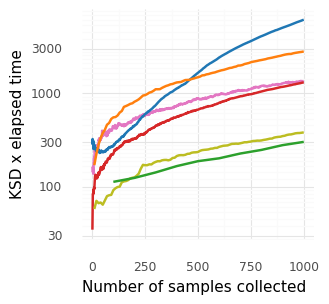}
	\includegraphics[width=0.48\textwidth,height=0.44\textwidth]{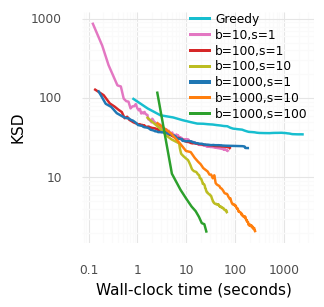}
	\includegraphics[width=0.49\textwidth,height=0.44\textwidth]{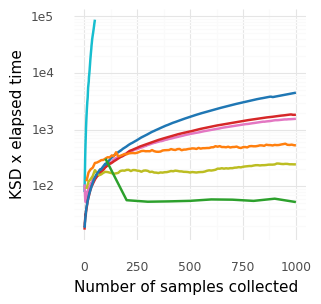}
	\caption{KSD  vs. wall-clock time, and KSD $\times$ time vs. number of selected samples, shown for the 38-dim calcium signalling model (top two panes) and 4-dim Lotka--Volterra model (bottom two panes). The kernel length-scale in each case was set using the median heuristic \citep{garreau2017large}, and estimated in practice using a uniform subsample of 1000 points for each model. 
	The myopic algorithm of \cite{riabiz2020optimal} is included in the Lotka--Volterra plots---see main text for details.
	}
	 \label{fig: thinning}
	\end{minipage}
\vspace*{-1.8em}
\end{figure*}

\subsection{Bayesian Cubature} \label{subsec: BC}

\cite{larkin1972gaussian} and subsequent authors proposed to cast numerical cubature in the Bayesian framework, such that an integrand $f$ is \textit{a priori} modelled as a Gaussian process with covariance $k$, then conditioned on data $\mathcal{D} = \{f(x_i)\}_{i=1}^n$ as the integrand is evaluated.
The posterior standard deviation is \citep{briol2019probabilistic}
\begin{align} \textstyle
    \text{Std}\big[\int\! f \d \mu | \mathcal{D}\big] & = \!\!\!\!\!\! \min_{w_1,\dots,w_m \in \mathbb{R} }\!\!\! \textstyle \text{MMD}_{\mu,k}\Big(\sum\limits_{i=1}^m w_i \delta(x_i)\Big) . \label{eq: post var}
\end{align}
The selection of $x_i$ to minimise \eqref{eq: post var} is impractical since evaluation of \eqref{eq: post var} has complexity $O(m^3)$.
\cite{huszar2012optimally} and \cite{briol2015frank} noted that \eqref{eq: post var} can be bounded above by fixing $w_i = \frac{1}{m}$, and that quantisation methods give a practical means to minimise this bound.
All results in \Cref{sec: theory} can therefore be applied and, moreover, the bound is expected to be quite tight---$w_i \ll \frac{1}{m}$ implies that $x_i$ was not optimally placed, thus for optimal $x_i$ we anticipate $w_i \approx \frac{1}{m}$. 

BC is most often used when evaluation of $f$ has a high computational cost, and one is prepared to expend resources in the optimisation of the point set $\{x_i\}_{i=1}^m$.
Our focus in this section is therefore on the quality of the point set obtained, irrespective of computational cost. 
Figure \ref{fig: visual example} suggested that the approximation quality of non-myopic methods depends on $s$. 
Figure \ref{fig: vary batch} compares the selection of 60 from 1000 independently sampled points from a mixture of 20 Gaussians, varying $s$. This gives a set of step functions. 
Less myopic selections are seen to outperform more myopic ones. Note in particular that MMD of the myopic method ($s=1$) is observed to decrease non-monotonically. This is a manifestation of the phenomenon also seen in Figure \ref{fig: visual example}, where a particular selection may temporarily worsen the quality of the overall approximation.

Next we consider applications in Bayesian statistics, where both approximation quality and computation time are important. 
In what follows the density $p_\mu$ will be available only up to an unknown normalisation constant and thus KSD---which requires only that $u_\mu = \nabla \log p_\mu$ can be evaluated---will be used.

\subsection{Thinning of Markov Chain Output} \label{subsec: MCMC}
The use of quantisation to `thin' Markov chain output was proposed in \cite{riabiz2020optimal}, who studied greedy myopic algorithms based on KSD.
We revisit the applications from that work to determine whether our methods offer a performance improvement.
Unlike \Cref{subsec: BC}, the cost of our algorithms must now be assessed, since their runtime may be comparable to the time required to produce Markov chain output itself.
The datasets\footnote{Available at \url{https://doi.org/10.7910/DVN/MDKNWM}.} consist of (i) $4\times 10^6$ samples from the 38-parameter intracellular calcium signalling model of \cite{hinch2004simplified}, and (ii) $2\times 10^6$ samples from a 4-parameter Lotka--Volterra predator-prey model. The MCMC chains are both highly auto-correlated and start far from any mode. 
The greedy KSD approach of \cite{riabiz2020optimal} was found to slow down dramatically after selecting $10^3$ samples, due to the need to compute the kernel between selected points and \emph{all} points in the candidate set at each iteration. 
We employ mini-batching ($b < n$) to ameliorate this, and also investigate the effectiveness of non-myopic selection.

Figure \ref{fig: thinning} plots KSD against time, with the number of collected samples $m$ fixed at 1000, as well as time-adjusted KSD against $m$ for both models. 
This acknowledges that both approximation quality and computational time are important. 
In both cases, larger mini-batches were able to perform better \textit{provided} that $s$ was large enough to realise their potential. Non-myopic selection shows a significant improvement over batch-myopic in the Lotka--Volterra model, and a less significant (though still visible) improvement in the calcium signalling model. The practical upper limit on $b$ for non-myopic methods (due to the requirement to optimise over all $b$ points) may make performance for larger $s$ poorer relatively; in a larger and more complex model, there may be fewer than $s$ `good' samples to choose from given moderate $b$. 
This suggests that control of the ratio $s/b$ may be important; the best results we observed occurred when $s/b = 10^{-1}$.

A comparison to the original myopic algorithm (i.e. $b=n$) of \cite{riabiz2020optimal} is incuded for the Lotka--Volterra model. 
This is implemented using the same code and machine as the other simulations. The cyan line shown in the bottom two panes of Figure \ref{fig: thinning} represents only 50 points (not 1000); collecting just these took 42 minutes. 
This algorithm is slower still for the calcium signalling model, so it was omitted.

\subsection{Comparison with Previous Approaches} \label{sec: continuous comparison}

\begin{figure}[t!]
\centering
\vspace*{-0.5em}
\hspace*{-1.2em}
\includegraphics[width=0.47\textwidth,height=0.36\textwidth]{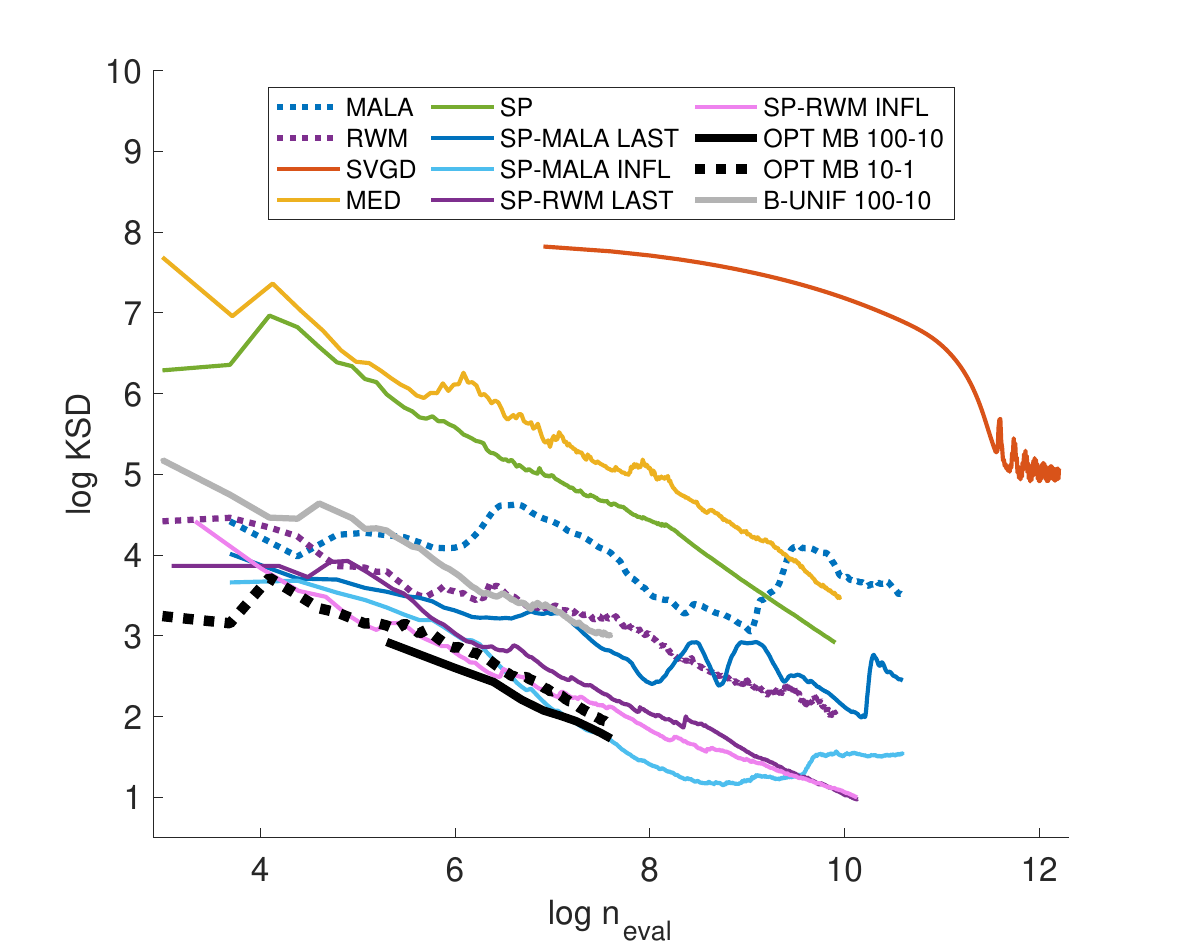}
	\caption{Comparison of quality of approximation for various methods, adjusted by the number of evaluations $n_\mathrm{{eval}}$ of $\mu$ or $u_\mu$. For details of  line labels, refer to the main text, and to Fig. 4 of \cite{chen2019stein}.}\label{fig: chen comparison}
	\vspace*{-0.0em}
\end{figure}

Here we compare against approaches based on continuous optimisation, reproducing a $10$-dimensional ODE inference task due to \citet{chen2019stein}. 
The aim is to minimise KSD whilst controlling the number of evaluations $n_{\text{eval}}$ of either the (un-normalised) target $\mu$ or its log-gradient $u_\mu$.
\Cref{fig: chen comparison} reports results for \textit{random walk Metropolis} (RWM), the \textit{Metropolis-adjusted Langevin algorithm} (MALA), \textit{Stein variational gradient descent} (SVGD), \textit{minimum energy designs} (MED), \textit{Stein points} (SP), and \textit{Stein point MCMC}  \citep[four flavours, denoted SP-$\ast$, described in][]{chen2019stein}. The method from Algorithm 3, shown as a black solid line (OPT MB 100-10; $b = 100$, $s = 10$) and dashed line (OPT MB 10-1; $b = 10$, $s = 1$), was applied to select $100$ states from the first $mb$ states visited in the RWM sample path, with $m$ increasing and $b$ fixed.
The resulting quantisations are competitive with those produced by existing methods, at comparable computational cost. 
We additionally include a comparison with batch-uniform selection (B-UNIF 100-10; $b=100$, $s=10$, drawn uniformly from the RWM output) in light grey.

\section{Theoretical Assessment} \label{sec: theory}

This section presents a theoretical assessment of Algorithms \ref{alg: myopic mmd}--\ref{alg: non-myopic mmd minibatch}.
Once stated, a standing assumption is understood to hold for the remainder of the main text.

\begin{standing}
Let $\mathcal{X}$ be a measurable space equipped with a probability measure $\mu$.
Let $k:\mathcal{X}\times\mathcal{X}\rightarrow\mathbb{R}$ be symmetric positive definite and satisfy $C_{\mu,k}^2 := \iint\! k(x,y) \mathrm{d}\mu(x) \mathrm{d}\mu(y) < \infty$. 
\end{standing}

For the kernels $k_\mu$ described in \Cref{subsec: KSD}, $C_{\mu,k_\mu} = 0$ and the assumption is trivially satisfied.
Our first result is a finite-sample-size error bound for non-myopic algorithms when the candidate set is fixed:

\begin{theorem} \label{thm: nonmyopic fixed}
Let $\{x_i\}_{i=1}^n \subset \mathcal{X}$ be fixed and let $C_{n,k}^2 := \max_{i=1,\dots,n} k(x_i,x_i)$. 
Consider an index sequence $\pi$ of length $m$ and with selection size $s$ produced by \Cref{alg: non-myopic mmd}. 
Then for all $m \geq 1$ it holds that 
\begin{align*} 
	& \textstyle \mathrm{MMD}_{\mu,k}\left(\frac{1}{ms} \sum_{i=1}^m \sum_{j=1}^s \delta(x_{\pi(i,j)}) \right)^2 \\ 
	& \hspace{8pt}   \leq \textstyle \underset{\substack{1^\top\!w = 1 \\w_i\geq 0}}{\mathrm{min}}\ \mathrm{MMD}_{\mu,k} \big( \sum_{i=1}^n  w_i\delta(x_i) \big)^2 + C^2 \left(\frac{1+\log m}{m}\right) ,
\end{align*}
with $C := C_{\mu,k} + C_{n,k}$ an $m$-independent constant.
\end{theorem}

The proof is provided in \Cref{app: proof of nonmyopic fixed}.
Aside from providing an explicit error bound, we see that the output of \Cref{alg: non-myopic mmd} converges in MMD to the optimal (weighted) quantisation of $\mu$ that is achievable using the candidate point set.
Interestingly, all bounds we present are independent of $s$.

\begin{remark}
Theorems \ref{thm: nonmyopic fixed}--\ref{thm: sampled 2} are stated for general MMD and apply in particular to KSD, for which we set $k = k_\mu$.
These results extend the work of \cite{chen2018stein,chen2019stein} and \cite{riabiz2020optimal}, who considered only myopic algorithms (i.e., $s=1$).
\end{remark}

\begin{remark}
	The theoretical bounds being independent of $s$ do not necessarily imply that $s = 1$ is optimal in practice; indeed our empirical results in Section \ref{sec: empirical} suggest that it is not.
\end{remark}

Our remaining results explore the cases where the candidate points are randomly sampled.
Independent and dependent sampling is considered and, in each case, the following moment bound will be assumed:

\begin{standing}
For some $\gamma > 0$, 	$C_1 := \textstyle \sup_{i \in \mathbb{N}}
\mathbb{E} \left[ e^{\gamma k(x_i,x_i)} \right] < \infty$, where the
expectation is taken over $x_1,\dots,x_n$.
\end{standing}
In the first randomised setting, the $x_i$ are independently sampled from $\mu$, as would typically be possible when $\mu$ is explicit:

\begin{theorem} \label{thm: sampled 1}
Let $\{x_i\}_{i=1}^n \subset \mathcal{X}$ be independently sampled from $\mu$. 
Consider an index sequence $\pi$ of length $m$ produced by \Cref{alg: non-myopic mmd}. 
Then for all $s\in \mathbb{N}$ and all $m,n \geq 1$ it holds that
\begin{align*} 
	& \hspace{0pt} \textstyle \mathbb{E} \Big[ \mathrm{MMD}_{\mu,k}\left(\frac{1}{ms} \sum_{i=1}^m \sum_{j=1}^s \delta(x_{\pi(i,j)}) \right)^2 \Big] \\ 
	& \hspace{30pt} \leq \textstyle \frac{\log(C_1)}{n \gamma}  + 2 \left( C_{\mu,k}^2 + \frac{\log(n C_1)}{\gamma} \right) \left(\frac{1+\log m}{m}\right) .
\end{align*}
\end{theorem}
\vspace{-0.8em}
The proof is provided in \Cref{app: proof of sampled 1}.
It is seen that $n$ must be asymptotically increased with $m$ in order for the approximation provided by \Cref{alg: non-myopic mmd} to be consistent.
No smoothness assumptions were placed on $k$; such assumptions can be used to improve the $O(n^{-1})$ term \citep[as in Thm. 1 of][]{ehler2019optimal}, but we did not consider this useful since the $O(m^{-1})$ term is the bottleneck in the bound.

An analogous (but more technically involved) argument leads to a finite-sample-size error bound when mini-batches are used:

\begin{theorem} \label{thm: mini batch}
Let each mini-batch $\{x_j^i\}_{j=1}^b \subset \mathcal{X}$ be independently sampled from $\mu$. 
Consider an index sequence $\pi$ of length $m$ produced by \Cref{alg: non-myopic mmd minibatch}. 
Then $\forall\ m,n \geq 1$
\begin{align*} 
	& \hspace{0pt} \textstyle \mathbb{E} \Big[ \mathrm{MMD}_{\mu,k}\left(\frac{1}{ms} \sum_{i=1}^m \sum_{j=1}^s \delta(x_{\pi(i,j)}^i) \right)^2 \Big] \\ 
	& \hspace{30pt} \leq \textstyle \frac{\log(C_1)}{b \gamma}  + 2 \left( C_{\mu,k}^2 + \frac{\log(b C_1)}{\gamma} \right) \left(\frac{1+\log m}{m}\right) .
\end{align*}
\end{theorem}
\vspace{-0.8em}
The proof is provided in \Cref{app: proof of mini batch}.
The mini-batch size $b$ plays an analogous role to $n$ in \Cref{thm: sampled 1} and must be asymptotically increased with $m$ in order for \Cref{alg: non-myopic mmd minibatch} to be consistent. 

In our second randomised setting the candidate set arises as a Markov chain sample path.
	Let $V$ be a function $V: \mathcal{X} \to [1,\infty)$ and, for a function $f : \mathcal{X} \rightarrow \mathbb{R}$ and a measure $\mu$ on $\mathcal{X}$, let 
\begin{equation*} \textstyle
	\|f\|_V := \sup_{x \in \mathcal{X}} \frac{|f(x)|}{V(x)}\,, \;\;\; \|\mu\|_V := \sup_{\|f\|_V \leq 1} \left| \int f \mathrm{d}\mu \right|.
\end{equation*}
	A $\psi$-irreducible and aperiodic Markov chain with $n{\text{th}}$ step transition kernel $\mathrm{P}^n$ is \emph{$V$\!-uniformly ergodic} if and only if there exists $R \in [0,\infty)$ and $ \rho \in (0,1)$ such that 
	\begin{align}
	\| \mathrm{P}^n(x,\cdot)-P \|_{V} \leq R V(x) \rho^n \label{eq: V unif ergod}
	\end{align}
	for all initial states $x \in \mathcal{X}$ and all $n \in \mathbb{N}$ \citep[see Thm. 16.0.1 of][]{Meyn2012}.

	\begin{theorem} \label{thm: sampled 2}
		Assume that $\int k(x,\cdot) \mathrm{d}\mu(x) = 0$ for all $x \in \mathcal{X}$.
		Consider a $\mu$-invariant, time-homogeneous, reversible Markov chain $\{x_i\}_{i \in \mathbb{N}} \subset \mathcal{X}$ generated using a $V$\!-uniformly ergodic transition kernel, such that \eqref{eq: V unif ergod} is satisfied with $V(x) \geq \sqrt{k(x,x)}$ for all $x \in \mathcal{X}$.
		Suppose that $C_2 := \sup_{i \in \mathbb{N}} \mathbb{E}[\sqrt{k(x_i,x_i)} V(x_i) ] < \infty$.
		Consider an index sequence $\pi$ of length $m$ and selection subset size $s$ produced by \Cref{alg: non-myopic mmd}. 
		Then, with $C_3 = \frac{2 R \rho}{1-\rho}$, we have that
		\begin{align*}
	& \textstyle \mathbb{E} \Big[ \mathrm{MMD}_{\mu,k}\left(\frac{1}{ms} \sum_{i=1}^m \sum_{j=1}^s \delta(x_{\pi(i,j)}) \right)^2 \Big] \\ 
	& \hspace{15pt} \leq \textstyle \frac{\log(C_1)}{n \gamma} + \frac{C_2 C_3}{n}  + 2 \left( C_{\mu,k}^2 + \frac{\log(n C_1)}{\gamma} \right) \left(\frac{1+\log m}{m}\right).
		\end{align*}
	\end{theorem}
	
The proof is provided in \Cref{app: proof of sampled 2}.
Analysis of mini-batching in the dependent sampling context appears to be more challenging and was not attempted.

\section{Discussion} \label{sec: discuss}

This paper focused on quantisation using MMD, proposing and analysing novel algorithms for this task, but other integral probability metrics could be considered.
More generally, if one is interested in compression by means other than quantisation then other approaches may be useful, such as Gaussian mixture models and related approaches from the literature on density estimation \citep{silverman1986density}.

Some avenues for further research include:
(i) extending symmetric structure in $\mu$ to the set of representative points \citep{karvonen2018symmetry}; 
(ii) characterising an \textit{optimal} sampling distribution from which elements of the candidate set can be obtained \citep{bach2017equivalence};
(iii) further applications of our method, for example to Bayesian neural networks, where quantisation of the posterior provides a promising route to reduce the cost of predicting each label in the test dataset. \hfill \textbullet

\ifnum\statePaper=1
    {
    \acknowledgments{
    
    The authors are grateful to Lester Mackey and Luc Pronzato for helpful feedback on an earlier draft, and to Wilson Chen for sharing the code used to produce Figure \ref{fig: chen comparison}.
    This work was supported by the Lloyd's Register Foundation programme on data-centric engineering at the Alan Turing Institute, UK.
    MR was supported by the British
    Heart Foundation--Alan Turing Institute cardiovascular data science award (BHF;~SP/18/6/33805).
    }
    }
\fi

\nocite{gurobi}
\nocite{mosek}

\bibliographystyle{apalike}
\balance
\interlinepenalty=10000
\bibliography{bibliography}

\newpage
\onecolumn
\appendix
\section*{Supplementary Material}

This supplement is structured as follows:
In \Cref{sec: proofs} we present proofs for all novel theoretical results stated in \Cref{sec: theory} of the main text. In Appendices \ref{appendix: sdr} and \ref{sec: extra experiments} we provide additional experimental results to support the discussion in \Cref{sec: empirical} of the main text.

\section{Proof of Theoretical Results} \label{sec: proofs}

In what follows we let $\mathcal{H}$ denote the reproducing kernel Hilbert space $\mathcal{H}(k)$ reproduced by the kernel $k$ and let $\|\cdot\|_{\mathcal{H}}$ denote the induced norm in $\mathcal{H}$.

\subsection{Proof of \Cref{thm: nonmyopic fixed}} \label{app: proof of nonmyopic fixed}

	To start the proof, define 
	\[
	\begin{aligned}
	a_m & := (ms)^2 \; \mathrm{MMD}_{\mu,k}\left(\frac{1}{ms} \sum_{i=1}^m \sum_{j=1}^s \delta(x_{\pi(i,j)}) \right)^2 \\
	&= \sum_{i=1}^m\sum_{i'=1}^m\sum_{j=1}^s\sum_{j'=1}^s k(x_{\pi(i,j)},x_{\pi(i',j')}) - 2ms\sum_{i=1}^m\sum_{j=1}^s\int k(x_{\pi(i,j)},x)\d\mu(x) + (ms)^2\iint k(x,x')\d\mu(x)\d\mu(x') \\
	f_m(\cdot)  & := \sum_{i=1}^m\sum_{j=1}^s k(x_{\pi(i,j)},\cdot) - ms\int k(\cdot,x)\d\mu(x)\\
	\end{aligned}
	\]
	and note immediately that $a_m = \lVert f_m \rVert^2_\mathcal{H}$. Then we can write a recursive relation
	\[
	\begin{aligned}
		a_m &= a_{m-1} + \underbracket{\vphantom{\sum_{j=1}}\sum_{j=1}^s\sum_{j'=1}^s k(x_{\pi(m,j)},x_{\pi(m,j')}) + 2\sum_{i=1}^{m-1}\sum_{j=1}^s\sum_{j'=1}^sk(x_{\pi(m,j)},x_{\pi(i,j')})  - \vphantom{\sum_{j=1}}2ms\sum_{j=1}^s\int k(x_{\pi(m,j)},x)\d \mu(x)}_{(*)} \\ &  \qquad\qquad\qquad\qquad\qquad  \underbracket{-\, 2s\sum_{i=1}^{m-1}\sum_{j=1}^{s}\int k(x_{\pi(i,j)},x)\d\mu(x)\,+\, \vphantom{\sum_{j=1}}s^2(2m-1)\iint k(x,x')\d\mu(x)\d\mu(x')}_{(**)} \\
	\end{aligned}
	\]
We will first derive an upper bound for $(*)$, then one for $(**)$. 

\paragraph{Bounding $(*)$:}
Noting that the algorithm chooses the $S \in \{1,\dots,n\}^s$ that minimises 
\begin{multline*}
\sum_{j\in S}\sum_{j' \in S} k(x_{j},x_{j'}) + 2\sum_{j\in S}\sum_{j'=1}^s \sum_{i=1}^{m-1} k(x_{j},x_{\pi(i,j')}) - 2ms\sum_{j \in S} \int k(x_j,x) \d\mu(x) \\
=  \sum_{j\in S}\sum_{j' \in S} k(x_{j},x_{j'})  - 2s\sum_{j\in S}\int k(x_j,x)\d\mu(x) + 2\sum_{j\in S}f_{m-1}(x_j) ,
\end{multline*}
we therefore have that 
\begin{align}
(*) &= \min_{S \in \{1,\dots,n\}^s} \left[\sum_{j\in S}\sum_{j' \in S} k(x_{j},x_{j'})  - 2s\sum_{j\in S}\int k(x_j,x)\d\mu(x) + 2\sum_{j\in S}f_{m-1}(x_j)\right] \nonumber \\
&\leq \max_{S \in \{1,\dots,n\}^s}  \left[ \sum_{j\in S}\sum_{j' \in S} k(x_{j},x_{j'})  - 2s\sum_{j\in S}\int k(x_j,x)\d\mu(x)  \right] + 2 \min_{S \in \{1,\dots,n\}^s} \sum_{j\in S}  f_{m-1}(x_j) \nonumber \\
& = \max_{S \in \{1,\dots,n\}^s} \left[  \sum_{j\in S}\sum_{j' \in S} k(x_{j},x_{j'}) - 2 s \sum_{j\in S} \int  \vphantom{X^{X^X}}\left\langle \vphantom{X^X} k(x_j,\cdot),k(x,\cdot) \right\rangle_\mathcal{H}  \d\mu(x) \right] + 2 \min_{S \in \{1,\dots,n\}^s} \sum_{j\in S}f_{m-1}(x_j) \label{eq: RP0} \\
&\leq \max_{S \in \{1,\dots,n\}^s}  \left[  \sum_{j\in S}\sum_{j' \in S} k(x_{j},x_{j'}) + 2s  \sum_{j\in S}\left\Vert  \vphantom{X_X^X} k(x_j,\cdot) \right\Vert_\mathcal{H} \cdot \int \left\Vert \vphantom{X_X^X} k(x,\cdot) \right\Vert_\mathcal{H}  \d\mu(x) \right] + 2 \min_{S \in \{1,\dots,n\}^s} \sum_{j\in S} f_{m-1}(x_j) \label{eq: CS0} \\
&\leq s^2 \max_{j \in \{1,\dots,n\}} k(x_{j},x_{j}) + 2s^2 \max_{j \in \{1,\dots,n\}} \sqrt{ k(x_{j},x_{j})} \cdot \int\!\sqrt{k(x,x)}\,\d\mu(x)  + 2 \min_{S \in \{1,\dots,n\}^s}\sum_{j\in S} f_{m-1}(x_j) \nonumber \\ 
& \leq s^2 C_{n,k}^2 + 2s^2 C_{n,k} \left( \int k(x,x) \mathrm{d}\mu(x) \right)^{1/2}  + 2 \min_{S \in \{1,\dots,n\}^s}\sum_{j\in S} f_{m-1}(x_j) \label{eq: Jensen0} \\ 
&= s^2C_{n,k}^2 +  2s^2C_{n,k}C_{\mu,k}  + 2 \min_{S \in \{1,\dots,n\}^s}\sum_{j\in S} f_{m-1}(x_j) \label{eq: last bit}
\end{align}
In \eqref{eq: RP0} we used the reproducing property, while in \eqref{eq: CS0} we used the Cauchy--Schwarz inequality and in \eqref{eq: Jensen0} we used Jensen's inequality.
To bound the third term in \eqref{eq: last bit}, we write 
\[
\min_{S \in \{1,\dots,n\}^s}\sum_{j\in S} f_{m-1}(x_j) = \min_{S \in \{1,\dots,n\}^s} \left\langle f_{m-1},\sum_{j\in S}k(\cdot,x_j) \right\rangle_{\!\!\mathcal{H}}
\]
Define $\mathcal{M}$ as the convex hull in $\mathcal{H}$ of $\left\{s^{-1}\sum_{j \in S} k(\cdot,x_j), S \in \{1,\dots,n\}^s\right\}$. 
Since the extreme points of $\mathcal{M}$ correspond to the vertices $(x_i,\dots,x_i)$ we have that 
\[
\mathcal{M} = \left\{\sum_{i=1}^n c_i k(\cdot,x_i) : c_i\geq 0, \sum_{i=1}^n c_i  = 1\right\} .
\]
Then we have, for any $h \in \mathcal{M}$,
\[
\langle f_{m-1}, h \rangle_\mathcal{H} = \left\langle f_{m-1}, \sum_{i=1}^n c_i k(\cdot,x_i) \right\rangle_{\!\!\mathcal{H}} =  \sum_{i=1}^n c_i f_{m-1}(x_i) .
\]
This linear combination is clearly minimised by taking each of the $x_i$ equal to a candidate point $x_j$ that minimises $f_{m-1}(x_j)$, and taking the corresponding $c_j = 1$, and all other $c_i = 0$. 
Now consider an element $h_w  = \sum_{i=1}^n w_i k(\cdot,x_i)$ for which the weights $w = (w_1,\dots,w_n)^\top$ minimise $\text{MMD}_{\mu,k}(\sum_{i=1}^n w_i \delta(x_i))$ subject to $1^\top w = 1$ and $w_i \geq 0$. 
Clearly $h_w \in \mathcal{M}$. Thus
\[
\min_{S \in \{1,\dots,n\}^s}\sum_{j\in S} f_{m-1}(x_j)  = s\cdot \inf_{h \in \mathcal{M}}\langle f_{m-1},h\rangle_\mathcal{H} \leq s\cdot \langle f_{m-1},h_w\rangle_\mathcal{H} .
\]
Combining this with \eqref{eq: last bit} provides an overall bound on $(*)$.

\paragraph{Bounding $(**)$:}
To upper bound $(**)$ we can in fact just use an equality;
\[
\begin{aligned}
(**) &= -2s\left[ \sum_{i=1}^{m-1}\sum_{j=1}^{s}\int k(x_{\pi(i,j)},x)\d\mu(x)\,+\, \vphantom{\sum_{j=1}}s(m-1)\iint k(x,x')\d\mu(x)\d\mu(x') \right] \\
& \qquad\qquad\qquad\qquad\qquad\qquad\qquad\qquad\qquad\qquad\qquad\qquad\qquad + s^2\iint k(x,x')\d\mu(x)\d\mu(x') \\
&= -2s \langle f_{m-1} , h_\mu \rangle_\mathcal{H} + s^2 \lVert  h_\mu \rVert_\mathcal{H}^{2}
\end{aligned}
\]
where $h_\mu = \int k(\cdot,x) \d\mu(x)$. 

\paragraph{Bound on the Iterates:}
Combining our bounds on $(*)$ and $(**)$, we obtain
\[
\begin{aligned}
a_m &\leq a_{m-1}+ s^2C_{n,k}^2 + 2s^2C_{n,k}C_{\mu,k} + 2 s\langle f_{m-1},h_w\rangle_\mathcal{H} -2s \langle f_{m-1} , h_\mu \rangle_\mathcal{H} + s^2 \lVert  h_\mu \rVert_\mathcal{H}^2 \\ 
& =a_{m-1}+ s^2C_{n,k}^2 + 2s^2C_{n,k}C_{\mu,k} + 2s \langle f_{m-1},h_w- h_\mu \rangle_\mathcal{H}+ s^2\lVert  h_\mu \rVert_\mathcal{H}^2\\
&\leq  a_{m-1}+ s^2C_{n,k}^2 + 2s^2C_{n,k}C_{\mu,k}  + 2 s\lVert f_{m-1}\rVert_\mathcal{H}\cdot \lVert h_w - h_\mu \rVert_\mathcal{H} + s^2\lVert  h_\mu \rVert_\mathcal{H}^2 \\ 
&\leq  a_{m-1}+ \left(s^2C_{n,k}^2 + 2s^2C_{n,k}C_{\mu,k} + s^2C_{\mu,k}^2\right) + 2 s\sqrt{a_{m-1}} \cdot \lVert  h_w - h_\mu \rVert_\mathcal{H}
\end{aligned}
\]
The last line arises because 
\begin{align}
\|h_\mu\|^2_\mathcal{H} = \iint k(x,x') \d\mu(x) \d\mu(x') & = \iint \langle k(x,\cdot) , k(x',\cdot) \rangle \d\mu(x) \d\mu(x') \label{eq: RP20} \\
& \leq \iint | \langle k(x,\cdot) , k(x',\cdot) \rangle | \d\mu(x) \d\mu(x') \nonumber \\
& \leq \iint \| k(x,\cdot) \|_{\mathcal{H}} \| k(x',\cdot) \|_{\mathcal{H}} \d\mu(x) \d\mu(x') \label{eq: CS20} \\
& = \left( \int \sqrt{k(x,x)} \d\mu(x) \right)^2 \nonumber \\
& \leq \int k(x,x) \d \mu(x) = C_{\mu,k}^2 . \label{eq: Jensen20}
\end{align}
In \eqref{eq: RP20} we used the reproducing property, while in \eqref{eq: CS20} we used the Cauchy--Schwarz inequality and in \eqref{eq: Jensen20} we used Jensen's inequality.

We now note that
\[
\begin{aligned}
\lVert  h_w - h_\mu \rVert_\mathcal{H}^2 &= \langle h_w - h_\mu , h_w - h_\mu \rangle_\mathcal{H} \\
&= \left\langle \sum_{i=1}^n w_i k(\cdot,x_i) - \int k(\cdot,x) \d\mu(x) , \sum_{i'=1}^n w_{i'} k(\cdot,x_{i'}) - \int k(\cdot,x') \d\mu(x') \right\rangle_\mathcal{\!\!H} \\
&= \sum_{i=1}^n \sum_{i'=1}^n w_i w_{i'} k(x_i,x_{i'}) - 2 \sum_{i=1}^n w_i \int k(x_i, x) \d\mu(x) + \iint k(x,x') \d\mu(x) \d\mu(x') \\
&= \mathrm{MMD}_{\mu,k} \left( \sum_{i=1}^n  w_i\delta(x_i) \right)^2 \; =: \Phi^2 ,
\end{aligned}
\]
which gives
\[
a_m \leq a_{m-1}+  s^2(C_{n,k} + C_{\mu,k})^2 + 2s \sqrt{a_{m-1}}\cdot\Phi
\]
as an overall bound on the iterates $a_m$.

\paragraph{Inductive Argument:}
Next we follow a similar argument to Theorem 1 in \cite{riabiz2020optimal} to establish an induction in $a_m$. 
Defining $C^2 := (C_{n,k} + C_{\mu,k})^2$ for brevity and noting that $C^2$ is a constant satisfying $C^2 \geq 0$, we assert
\[
a_m \leq (sm)^2 (\Phi^2 + K_m), \qquad \text{with} \qquad K_m := \frac{1}{m} (C^2 - \Phi^2)\sum_{j=1}^m \frac{1}{j}
\]
For $m=1$, we have $a_1 \leq s^2(C_{n,k}^2 + 2C_{n,k}C_{\mu,k} + C_{\mu,k}^2) = s^2 C^2$, so the root of the induction holds. 
We now assume that $a_{m-1} \leq s^2(m-1)^2 (\Phi^2 + K_{m-1})$. Then
\begin{align}
a_m &\leq a_{m-1} + s^2C^2 +  2s \sqrt{a_{m-1}} \cdot \Phi \nonumber \\
&\leq s^2(m-1)^2 (\Phi^2 + K_{m-1}) + s^2C^2 +  2 s^2(m-1)\Phi\sqrt{\Phi^2 + K_{m-1}} \nonumber \\
&\leq s^2\left[ (m-1)^2 (\Phi^2 + K_{m-1}) + C^2 +  (m-1)(2\Phi^2 + K_{m-1})\right] \label{eq: algebra bit} \\
&= s^2\left[(m^2 - 1) \Phi^2 + m(m-1)K_{m-1}+ C^2 \right] \nonumber \\
&= s^2\bigg[ (m^2 - 1) \Phi^2 + m(C^2 - \Phi^2)\sum_{j=1}^{m-1}\frac{1}{j} + C^2\bigg] \nonumber \\
&= s^2\bigg[(m^2 - 1) \Phi^2 + m(C^2 - \Phi^2)\sum_{j=1}^{m}\frac{1}{j} - m(C^2 - \Phi^2)\frac{1}{m} + C^2  \bigg] \nonumber \\
&= s^2 \bigg[ m^2\Phi^2 + m(C^2 - \Phi^2)\sum_{j=1}^{m}\frac{1}{j}\bigg] \nonumber \\
&= (sm)^2(\Phi^2 + K_m), \nonumber
\end{align}
which proves the induction. 
Here \eqref{eq: algebra bit} follows from the fact that for any $a,b>0$, it holds that $2a\sqrt{a^2+b} \leq 2a^2 + b$. 

\paragraph{Overall Bound:}

To complete the proof, we first show that $\Phi^2 \leq C^2$ by writing
\[
\begin{aligned}
\Phi^2 = \|h_w - h_\mu\|^2_\mathcal{H}  \leq \|h_w\|^2_\mathcal{H} + 2\|h_w\|_\mathcal{H} \cdot\|h_\mu\|_\mathcal{H}  + \|h_\mu\|^2_\mathcal{H}
\end{aligned}
\]
and noting that, since $k(x_i,x_{i'}) \leq \sqrt{k(x_i,x_i)} \sqrt{k(x_{i'},x_{i'})}$ and $\sum_{i=1}^n w_i = 1$, it holds that
\[
\|h_w\|^2_\mathcal{H} = \sum_{i=1}^n \sum_{i'=1}^n w_i w_{i'} k(x_i,x_{i'}) \leq C_{n,k}^2 .
\]
We have already shown that $\|h_\mu\|^2 \leq C_{\mu,k}^2$, thus it follows that $\Phi^2 \leq C_{n,k}^2 + 2C_{n,k}C_{\mu,k} +C_{\mu,k}^2 \equiv C^2$ as required.

Using this bound in conjunction with the elementary series inequality $\sum_{j=1}^m j^{-1} \leq (1 + \log m)$, we have $K_m \geq 0$ and
\[
K_m =  \frac{1}{m}(C^2 - \Phi^2)\sum_{j=1}^{m}\frac{1}{j}\ \leq\   \frac{1}{m}C^2\sum_{j=1}^{m}\frac{1}{j}\ \leq \left(\frac{1+\log m}{m}\right)C^2
\]
Finally, the theorem follows by noting
\begin{align*}
\mathrm{MMD}_{\mu,k}\left(\frac{1}{ms} \sum_{i=1}^m \sum_{j=1}^s \delta(x_{\pi(i,j)}) \right)^2 = \frac{a_m}{(sm)^2} \leq \Phi^2 + K_m = \Phi^2 +\left(\frac{1+\log m}{m}\right)C^2 , 
\end{align*}
as claimed. \hfill $\square$

\paragraph{Remark:}
We observe that, in the myopic case only ($s=1$), one can alternatively recover \Cref{thm: nonmyopic fixed} as a consequence of Theorem 1 in \cite{riabiz2020optimal} \citep[refer also to Theorem 5 of][]{chen2019stein}. This can be seen by noting that $\text{MMD}_{\mu,k_0}(\nu) = \text{MMD}_{\mu,k}(\nu)$ for all $\nu \in \mathcal{P}(\mathcal{X})$, where $k_0$ is the kernel 
\begin{align}
k_0(x,y) := k(x,y) - \int k(x,x') \mathrm{d}\mu(x') - \int k(y,y') \mathrm{d}\mu(y') + \iint k(x',y') \mathrm{d}\mu(x') \mathrm{d}\mu(y'),  \label{eq: k0 kernel}
\end{align}
which satisfies the precondition $\displaystyle\int k_0(x,y') \mathrm{d}\mu(y') = 0$ for all $x \in \mathcal{X}$ in Theorem 1 of \cite{riabiz2020optimal}.
Indeed,
\begin{align*}
    \text{MMD}_{\mu,k_0}(\nu)^2 & = \left\| \int k_0(\cdot,y') \mathrm{d}\nu(y') - \int k_0(\cdot,y') \mathrm{d}\mu(y') \right\|_{\mathcal{H}(k_0)}^2 \\
    & = \left\| \int k_0(\cdot,y') \mathrm{d}\nu(y')  \right\|_{\mathcal{H}(k_0)}^2 \\
    & = \iint \left[ k(x,y) - \int k(x,y') \mathrm{d}\mu(y') - \int k(x',y) \mathrm{d}\mu(x') + \iint k(x',y') \mathrm{d}\mu(x') \mathrm{d}\mu(y') \right] \mathrm{d}\nu(x) \mathrm{d}\nu(y) \\
    & = \iint k(x,y) \mathrm{d}\mu(x) \mathrm{d}\mu(y) - \iint k(x,y) \mathrm{d}\mu(x) \mathrm{d}\nu(y) - \iint k(x,y) \mathrm{d}\nu(x) \mathrm{d}\nu(y) \\  & \hspace{10.5cm}+ \iint k(x,y) \mathrm{d}\nu(x) \mathrm{d}\nu(y) \\
    & = \text{MMD}_{\mu,k}(\nu)^2.
\end{align*}

\subsection{Proof of \Cref{thm: sampled 1}} \label{app: proof of sampled 1}

First note that the preconditions of \Cref{thm: nonmyopic fixed} are satisfied.
We may therefore take expectations of the bound obtained in \Cref{thm: nonmyopic fixed}, to obtain that:
\begin{align} 
\mathbb{E}\left[ \mathrm{MMD}_{\mu,k}\left(\frac{1}{ms} \sum_{i=1}^m \sum_{j=1}^s \delta(x_{\pi(i,j)}) \right)^2 \right] 
	&  \leq \mathbb{E}\left[ \underset{\substack{1^T\!w = 1 \\w_i\geq 0}}{\mathrm{min}}\ \mathrm{MMD}_{\mu,k} \left( \sum_{i=1}^n  w_i\delta(x_i) \right)^2 \right] + \mathbb{E}[C^2] \left(\frac{1+\log m}{m}\right) , \label{eq: expect bound}
\end{align}

To bound the first expectation we proceed as follows:
\begin{align}
\mathbb{E}\left[ \underset{\substack{1^T\!w = 1 \\w_i\geq 0}}{\mathrm{min}}\ \mathrm{MMD}_{\mu,k} \left( \sum_{i=1}^n  w_i\delta(x_i) \right)^2 \right] & \leq \mathbb{E}\left[  \mathrm{MMD}_{\mu,k} \left( \frac{1}{n} \sum_{i=1}^n  \delta(x_i) \right)^2 \right] \label{eq: opt mmd start} \\
& = \mathbb{E}\left[ \frac{1}{n^2} \sum_{i=1}^n\sum_{j=1}^n k(x_i,x_j) - \frac{2}{n}\sum_{i=1}^n\int k(x,x_i)\d \mu(x) + \iint k(x,y) \d \mu(x) \d \mu(y) \right] \nonumber \\
& = \mathbb{E}\left[ \frac{1}{n^2} \sum_{i=1}^n\sum_{j=1}^n k(x_i,x_j) \right] - \iint k(x,y) \d \mu(x) \d \mu(y)  \qquad \text{(since $x_i \sim \mu$)} \nonumber \\
& = \mathbb{E}\left[ \frac{1}{n^2} \sum_{i=1}^n k(x_i,x_i) + \frac{1}{n^2} \sum_{i=1}^n \sum_{j \neq i} k(x_i,x_j)  \right] - \iint k(x,y) \d \mu(x) \d \mu(y) \nonumber \\
& = \mathbb{E}\left[ \frac{1}{n^2} \sum_{i=1}^n k(x_i,x_i) \right] - \frac{1}{n} \iint k(x,y) \d \mu(x) \d \mu(y)  \qquad \text{(since $x_i \sim \mu$)} \nonumber \\
& = \frac{1}{n} \mathbb{E}\left[ k(x_1,x_1) \right] - \frac{C_{\mu,k}^2}{n} \nonumber \\
& = \frac{1}{n \gamma} \mathbb{E}\left[  \log e^{\gamma k(x_i,x_i)}  \right] - \frac{C_{\mu,k}^2}{n} \nonumber \\
& \leq \frac{1}{n \gamma} \log \mathbb{E}\left[ e^{\gamma k(x_i,x_i)}  \right] - \frac{C_{\mu,k}^2}{n} \nonumber \\
& \leq \frac{1}{n \gamma} \log(C_1) - \frac{C_{\mu,k}^2}{n} \nonumber \\
& \leq \frac{1}{n \gamma} \log(C_1) \label{eq: opt mmd end}  .
\end{align}

To bound the second expectation we use the fact that $C^2 = (C_{\mu,k} + C_{n,k})^2 \leq 2 C_{\mu,k}^2 + 2 C_{n,k}^2$ where $C_{\mu,k}$ is independent of the set $\{x_i\}_{i=1}^n$ to focus only on the term $C_{n,k}$.
Here we have that
\begin{align}
			\mathbb{E}[C_{n,k}^2] := \mathbb{E}\left[ \max_{i = 1,\dots, n} k(x_i, x_i) \right] & = \mathbb{E}\left[ \frac{1}{\gamma} \log \max_{i = 1, \dots, n} e^{ \gamma k(x_i, x_i) } \right] \label{eq: EC start} \\
			& \leq \mathbb{E}\left[ \frac{1}{\gamma} \log \sum_{i = 1}^{n} e^{ \gamma k(x_i, x_i) } \right] \nonumber \\
			& \leq \frac{1}{\gamma} \log \left( \sum_{i = 1}^{n} \mathbb{E}\left[ e^{\gamma k(x_i, x_i) } \right] \right) \; = \; \frac{\log(n C_1)}{\gamma} . \label{eq: EC end}
\end{align}

Thus we arrive at the overall bound
\begin{align*} 
\mathbb{E}\left[ \mathrm{MMD}_{\mu,k}\left(\frac{1}{ms} \sum_{i=1}^m \sum_{j=1}^s \delta(x_{\pi(i,j)}) \right)^2 \right] 
	&  \leq \frac{\log(C_1)}{n \gamma} + 2 \left( C_{\mu,k}^2 + \frac{\log(n C_1)}{\gamma} \right) \left(\frac{1+\log m}{m}\right) ,
\end{align*}
as claimed. \hfill $\square$

\paragraph{Remark:} We observe that, in the myopic case only ($s=1$), one can alternatively recover \Cref{thm: sampled 1} as a consequence of Theorem 2 in \cite{riabiz2020optimal}, once again using the observation that the kernel in \eqref{eq: k0 kernel} satisfies the preconditions of Theorem 2 in \cite{riabiz2020optimal}.

\subsection{Proof of Theorem \ref{thm: mini batch}} \label{app: proof of mini batch}

The following proof combines parts of the arguments used to establish \Cref{thm: nonmyopic fixed} and \Cref{thm: sampled 1}, with additional notation required to deal with the mini-batching involved.

In a natural extension to the proof of \Cref{thm: nonmyopic fixed}, we define 
	\[
	\begin{aligned}
	a_m & := (ms)^2 \; \mathrm{MMD}_{\mu,k}\left(\frac{1}{ms} \sum_{i=1}^m \sum_{j=1}^s \delta(x_{\pi(i,j)}^i) \right)^2 \\
	&= \sum_{i=1}^m\sum_{i'=1}^m\sum_{j=1}^s\sum_{j'=1}^s k(x_{\pi(i,j)}^i,x_{\pi(i',j')}^{i'}) - 2ms\sum_{i=1}^m\sum_{j=1}^s\int k(x_{\pi(i,j)}^i,x)\d\mu(x) + (ms)^2\iint k(x,x')\d\mu(x)\d\mu(x') \\
	f_m(\cdot)  & := \sum_{i=1}^m\sum_{j=1}^s k(x_{\pi(i,j)}^i,\cdot) - ms\int k(\cdot,x)\d\mu(x)\\
	\end{aligned}
	\]
	and note immediately that $a_m = \lVert f_m \rVert^2_\mathcal{H}$. Then, similarly to \Cref{thm: nonmyopic fixed}, we write a recursive relation
	\begin{align*}
		a_m &= a_{m-1} + \underbracket{\vphantom{\sum_{j=1}}\sum_{j=1}^s\sum_{j'=1}^s k(x_{\pi(m,j)}^m,x_{\pi(m,j')}^m) + 2\sum_{i=1}^{m-1}\sum_{j=1}^s\sum_{j'=1}^sk(x_{\pi(m,j)}^m,x_{\pi(i,j')}^i)  - 2ms\sum_{j=1}^s\int k(x_{\pi(m,j)}^m,x)\d \mu(x)}_{(*)} \\ &  \qquad\qquad\qquad\qquad\qquad  \underbracket{-\, 2s\sum_{i=1}^{m-1}\sum_{j=1}^{s}\int k(x_{\pi(i,j)}^i,x)\d\mu(x)\,+\, s^2(2m-1)\iint k(x,x')\d\mu(x)\d\mu(x')}_{(**)} .
	\end{align*}
We will first derive an upper bound for $(*)$, then one for $(**)$. 

\paragraph{Bounding $(*)$:}
Noting that at iteration $m$ the algorithm chooses the $S \in \{1,\dots,b\}^s$ that minimises 
\begin{multline*}
\sum_{j\in S}\sum_{j' \in S} k(x_{j}^m,x_{j'}^m) + 2\sum_{j\in S}\sum_{j'=1}^s \sum_{i=1}^{m-1} k(x_{j}^m,x_{\pi(i,j')}^i) - 2ms\sum_{j \in S} \int k(x_j^m,x) \d\mu(x) \\
=  \sum_{j\in S}\sum_{j' \in S} k(x_{j}^m,x_{j'}^m)  - 2s\sum_{j\in S}\int k(x_j^m,x)\d\mu(x) + 2\sum_{j\in S}f_{m-1}(x_j^m) ,
\end{multline*}
we have that 
\begin{align}
(*) &= \min_{S \in \{1,\dots,b\}^s} \left[\sum_{j\in S}\sum_{j' \in S} k(x_{j}^m,x_{j'}^m)  - 2s\sum_{j\in S}\int k(x_j^m,x)\d\mu(x) + 2\sum_{j\in S}f_{m-1}(x_j^m)\right] \nonumber \\
&\leq \max_{S \in \{1,\dots,b\}^s}  \left[ \sum_{j\in S}\sum_{j' \in S} k(x_{j}^m,x_{j'}^m)  - 2s\sum_{j\in S}\int k(x_j^m,x)\d\mu(x)  \right] + 2 \min_{S \in \{1,\dots,b\}^s} \sum_{j\in S}  f_{m-1}(x_j^m) \nonumber \\
& = \max_{S \in \{1,\dots,b\}^s} \left[  \sum_{j\in S}\sum_{j' \in S} k(x_{j}^m,x_{j'}^m) - 2 s \sum_{j\in S} \int \left\langle k(x_j^m,\cdot),k(x,\cdot) \right\rangle_\mathcal{H}  \d\mu(x) \right] + 2 \min_{S \in \{1,\dots,b\}^s} \sum_{j\in S}f_{m-1}(x_j^m) \label{eq: RP} \\
&\leq \max_{S \in \{1,\dots,n\}^b}  \left[  \sum_{j\in S}\sum_{j' \in S} k(x_{j}^m,x_{j'}^m) + 2s  \sum_{j\in S}\left\Vert  k(x_j^m,\cdot) \right\Vert_\mathcal{H} \cdot \int \left\Vert k(x,\cdot) \right\Vert_\mathcal{H}  \d\mu(x) \right] + 2 \min_{S \in \{1,\dots,b\}^s} \sum_{j\in S} f_{m-1}(x_j^m) \label{eq: CS} \\
&\leq s^2 \max_{j \in \{1,\dots,b\}} k(x_{j}^m,x_{j}^m) + 2s^2 \max_{j \in \{1,\dots,b\}} \sqrt{ k(x_{j}^m,x_{j}^m)} \cdot \int\!\sqrt{k(x,x)}\,\d\mu(x)  + 2 \min_{S \in \{1,\dots,b\}^s}\sum_{j\in S} f_{m-1}(x_j^m) \nonumber \\ 
& \leq s^2 C_{b,m,k}^2 + 2s^2 C_{b,m,k} \left( \int k(x,x) \mathrm{d}\mu(x) \right)^{1/2}  + 2 \min_{S \in \{1,\dots,b\}^s}\sum_{j\in S} f_{m-1}(x_j^m) \label{eq: Jensen} \\ 
&= s^2C_{b,m,k}^2 +  2s^2C_{b.m,k}C_{\mu,k}  + 2 \min_{S \in \{1,\dots,b\}^s}\sum_{j\in S} f_{m-1}(x_j^m) \nonumber
\end{align}
In \eqref{eq: RP} we used the reproducing property.
In \eqref{eq: CS} we used the Cauchy--Schwarz inequality.
In \eqref{eq: Jensen} we used Jensen's inequality.

To bound the third term, we write 
\[
\min_{S \in \{1,\dots,b\}^s}\sum_{j\in S} f_{m-1}(x_j^m) = \min_{S \in \{1,\dots,b\}^s} \left\langle f_{m-1},\sum_{j\in S}k(\cdot,x_j^m) \right\rangle_{\!\!\mathcal{H}}
\]
Define $\mathcal{M}_m$ as the convex hull in $\mathcal{H}$ of $\left\{s^{-1}\sum_{j \in S} k(\cdot,x_j^m), S \in \{1,\dots,b\}^s\right\}$. 
Since the extreme points of $\mathcal{M}_m$ correspond to the vertices $(x_i^m,\dots,x_i^m)$ we have that 
\[
\mathcal{M}_m = \left\{\sum_{i=1}^n c_i k(\cdot,x_i^m) : c_i\geq 0, \sum_{i=1}^n c_i  = 1\right\}
\]
Then we have for any $h \in \mathcal{M}_m$
\[
\langle f_{m-1}, h \rangle_\mathcal{H} = \left\langle f_{m-1}, \sum_{i=1}^n c_i k(\cdot,x_i^m) \right\rangle_{\!\!\mathcal{H}} =  \sum_{i=1}^n c_i f_{m-1}(x_i^m)
\]
This linear combination is clearly minimised by taking the $x_j^m \in \{x_i^m\}_{i=1}^b$ that minimises $f_{m-1}(x_j^m)$, and taking the corresponding $c_j = 1$, and all other $c_i = 0$. Now consider the element $h_w^m  = \sum_{i=1}^b w_i^m k(\cdot,x_i^m)$ for which the weights are equal to the optimal weight vector $w^m$. 
Clearly $h_w^m \in \mathcal{M}_m$. Thus
\[
\min_{S \in \{1,\dots,b\}^s}\sum_{j\in S} f_{m-1}(x_j^m)  = s\cdot \inf_{h \in \mathcal{M}_m}\langle f_{m-1},h\rangle_\mathcal{H} \leq s\cdot \langle f_{m-1},h_w^m\rangle_\mathcal{H} .
\]

\paragraph{Bounding $(**)$:}
Our bound on $(**)$ is actually just an equality:
\[
\begin{aligned}
(**) &= -2s\left[ \sum_{i=1}^{m-1}\sum_{j=1}^{s}\int k(x_{\pi(i,j)}^i,x)\d\mu(x)\,+\, \vphantom{\sum_{j=1}}s(m-1)\iint k(x,x')\d\mu(x)\d\mu(x') \right] \\
& \qquad\qquad\qquad\qquad\qquad\qquad\qquad\qquad\qquad\qquad\qquad\qquad\qquad + s^2\iint k(x,x')\d\mu(x)\d\mu(x') \\
&= -2s \langle f_{m-1} , h_\mu \rangle_\mathcal{H} + s^2 \lVert  h_\mu \rVert_\mathcal{H}^{2}
\end{aligned}
\]
where $h_\mu = \int k(\cdot,x) \d\mu(x)$. 

\paragraph{Bound on the Iterates:}
Combining our bounds on $(*)$ and $(**)$ leads to the following bound on the iterates:
\[
\begin{aligned}
a_m &\leq a_{m-1}+ s^2C_{b,m,k}^2 + 2s^2C_{b,m,k}C_{\mu,k} + 2 s\langle f_{m-1},h_w^m\rangle_\mathcal{H} -2s \langle f_{m-1} , h_\mu \rangle_\mathcal{H} + s^2 \lVert  h_\mu \rVert_\mathcal{H}^2 \\ 
& =a_{m-1}+ s^2C_{b,m,k}^2 + 2s^2C_{b,m,k}C_{\mu,k} + 2s \langle f_{m-1},h_w^m- h_\mu \rangle_\mathcal{H}+ s^2\lVert  h_\mu \rVert_\mathcal{H}^2\\
&\leq  a_{m-1}+ s^2C_{b,m,k}^2 + 2s^2C_{b,m,k}C_{\mu,k}  + 2 s\lVert f_{m-1}\rVert_\mathcal{H}\cdot \lVert h_w^m - h_\mu \rVert_\mathcal{H} + s^2\lVert  h_\mu \rVert_\mathcal{H}^2 \\ 
&\leq  a_{m-1}+ \left(s^2C_{b,m,k}^2 + 2s^2C_{b,m,k}C_{\mu,k} + s^2C_{\mu,k}^2\right) + 2 s\sqrt{a_{m-1}} \cdot \lVert  h_w^m - h_\mu \rVert_\mathcal{H}
\end{aligned}
\]
The last line arises because 
\begin{align}
\|h_\mu\|^2_\mathcal{H} = \iint k(x,x') \d\mu(x) \d\mu(x') & = \iint \langle k(x,\cdot) , k(x',\cdot) \rangle \d\mu(x) \d\mu(x') \label{eq: RP2} \\
& \leq \iint | \langle k(x,\cdot) , k(x',\cdot) \rangle | \d\mu(x) \d\mu(x') \nonumber \\
& \leq \iint \| k(x,\cdot) \|_{\mathcal{H}} \| k(x',\cdot) \|_{\mathcal{H}} \d\mu(x) \d\mu(x') \label{eq: CS2} \\
& = \left( \int \sqrt{k(x,x)} \d\mu(x) \right)^2 \nonumber \\
& \leq \int k(x,x) \d \mu(x) = C_{\mu,k}^2 \label{eq: Jensen2}
\end{align}
In \eqref{eq: RP2} we used the reproducing property.
In \eqref{eq: CS2} we used the Cauchy--Schwarz inequality.
In \eqref{eq: Jensen2} we used Jensen's inequality.

We now note that
\[
\begin{aligned}
\lVert  h_w^m - h_\mu \rVert_\mathcal{H}^2 &= \langle h_w^m - h_\mu , h_w^m - h_\mu \rangle_\mathcal{H} \\
&= \left\langle \sum_{i=1}^b w_i^m k(\cdot,x_i^m) - \int k(\cdot,x) \d\mu(x) , \sum_{i'=1}^b w_{i'}^m k(\cdot,x_{i'}^m) - \int k(\cdot,x') \d\mu(x') \right\rangle_\mathcal{\!\!H} \\
&= \sum_{i=1}^b \sum_{i'=1}^b w_i^m w_{i'}^m k(x_i^m,x_{i'}^m) - 2 \sum_{i=1}^b w_i^m \int k(x_i^m, x) \d\mu(x) + \iint k(x,x') \d\mu(x) \d\mu(x') \\
&= \mathrm{MMD}_{\mu,k} \left( \sum_{i=1}^b  w_i^m \delta(x_i^m) \right)^2 \; =: \Phi_m^2 ,
\end{aligned}
\]
which gives
\[
a_m \leq a_{m-1}+  s^2(C_{b,m,k} + C_{\mu,k})^2 + 2s \sqrt{a_{m-1}}\cdot\Phi_m .
\]

We then follow a similar argument to Theorem 1 in \cite{riabiz2020optimal} to establish an induction in $a_m$. 

\paragraph{Inductive Argument:}
Let $c_m^2 := (C_{b,m,k} + C_{\mu,k})^2$.
We assert
\[
\mathbb{E}[a_m] \leq (sm)^2 \mathbb{E}[ \Phi_m^2 + K_{m} ], \qquad \text{with} \qquad K_m := \frac{1}{m} (c_{m}^2 - \Phi_m^2)\sum_{j=1}^m \frac{1}{j}
\]
For $m=1$, the induction holds since $a_1 \leq s^2 c_1$. 
We now assume that $\mathbb{E}[a_{m-1}] \leq s^2(m-1)^2 \mathbb{E}[ \Phi_{m-1}^2 + K_{m-1} ]$. Then
\begin{align}
\mathbb{E}[a_m] &\leq \mathbb{E}[a_{m-1}] + s^2 \mathbb{E}[c_m^2] +  2s \mathbb{E}[\sqrt{a_{m-1}} \cdot \Phi_m] \nonumber \\
& = \mathbb{E}[a_{m-1}] + s^2 \mathbb{E}[c_m^2] +  2s \mathbb{E}[\sqrt{a_{m-1}}] \cdot \mathbb{E}[\Phi_m] \qquad \text{(independence of $a_{m-1}$ and $\Phi_m$)} \nonumber \\
& \leq \mathbb{E}[a_{m-1}] + s^2 \mathbb{E}[c_m^2] +  2s \sqrt{\mathbb{E}[a_{m-1}]} \cdot \mathbb{E}[\Phi_m] \qquad \text{(Jensen's inequality)} \nonumber \\
&\leq s^2(m-1)^2 \mathbb{E}[\Phi_{m-1}^2 + K_{m-1} ] + s^2 \mathbb{E}[c_m^2] +  2 s^2(m-1) \mathbb{E}[\Phi_m] \sqrt{ \mathbb{E}[\Phi_{m-1}^2 + K_{m-1} ]} \nonumber \\
&\leq s^2(m-1)^2 \mathbb{E}[\Phi_{m}^2 + K_{m-1}] + s^2 \mathbb{E}[c_m^2] +  2 s^2(m-1) \mathbb{E}[\Phi_{m}] \sqrt{ \mathbb{E}[\Phi_{m}^2 + K_{m-1}] } \qquad \text{(since $\Phi_{m-1} \stackrel{d}{=} \Phi_m$)} \nonumber \\
&\leq s^2(m-1)^2 \mathbb{E}[\Phi_{m}^2 + K_{m-1}] + s^2 \mathbb{E}[c_m^2] +  2 s^2(m-1) \mathbb{E}[\Phi_{m}^2]^{1/2} \sqrt{ \mathbb{E}[\Phi_{m}^2 + K_{m-1}] }  \qquad \text{(Jensen's inequality)} \nonumber \\
&\leq s^2\left[ (m-1)^2 \mathbb{E}[ \Phi_{m}^2 + K_{m-1} ] + \mathbb{E}[c_m^2] +  (m-1)(2\mathbb{E}[\Phi_{m}^2] + \mathbb{E}[K_{m-1}] )\right] \label{eq: algebra} \\
&= s^2 \mathbb{E}\left[(m^2 - 1) \Phi_{m}^2 + m(m-1) K_{m-1} + c_m^2 \right] \nonumber \\
&= s^2 \mathbb{E}\bigg[ (m^2 - 1) \Phi_{m}^2 + m(c_{m-1}^2 - \Phi_{m-1}^2)\sum_{j=1}^{m-1}\frac{1}{j} + c_m^2\bigg] \nonumber \\
&= s^2 \mathbb{E}\bigg[(m^2 - 1) \Phi_{m}^2 + m(c_{m-1}^2 - \Phi_{m-1}^2)\sum_{j=1}^{m}\frac{1}{j} - m(c_{m-1}^2 - \Phi_{m-1}^2)\frac{1}{m} + c_{m}^2  \bigg] \nonumber \\
&= s^2 \mathbb{E}\bigg[(m^2 - 1) \Phi_{m}^2 + m(c_{m-1}^2 - \Phi_{m-1}^2)\sum_{j=1}^{m}\frac{1}{j} - m(c_m^2 - \Phi_m^2)\frac{1}{m} + c_{m}^2  \bigg] \qquad \text{(since $c_{m-1} \stackrel{d}{=} c_m$, $\Phi_{m-1} \stackrel{d}{=} \Phi_m$)} \nonumber \\
&= s^2 \mathbb{E}\bigg[ m^2\Phi_{m}^2 + m(c_{m-1}^2 - \Phi_{m-1}^2)\sum_{j=1}^{m}\frac{1}{j} \bigg] \nonumber \\
&= (sm)^2\mathbb{E}[\Phi_{m}^2 + K_m] \nonumber
\end{align}
which proves the induction. 
The line \eqref{eq: algebra} follows from the second by the fact that for any $a,b>0$, it holds that $2a\sqrt{a^2+b} \leq 2a^2 + b$. 

\paragraph{Overall Bound:}
We now show that $\Phi_m^2 \leq c_m^2$, by writing
\[
\begin{aligned}
\Phi_m^2 = \|h_w^m - h_\mu\|^2_\mathcal{H}  \leq \|h_w^m\|^2_\mathcal{H} + 2\|h_w^m\|_\mathcal{H} \cdot\|h_\mu\|_\mathcal{H}  + \|h_\mu\|^2_\mathcal{H}
\end{aligned}
\]
and noting that since $\sum_{i=1}^n w_i^m = 1$, it holds that
\[
\|h_w^m\|^2_\mathcal{H} = \sum_{i=1}^b \sum_{i'=1}^b w_i^m w_{i'}^m k(x_i^m,x_{i'}^m) \leq C_{b,m,k}^2 .
\]
We have already shown that $\|h_\mu\|^2 \leq C_{\mu,k}^2$, thus it follows that $\Phi_m^2 \leq C_{b,m,k}^2 + 2C_{b,m,k}C_{\mu,k} +C_{\mu,k}^2 = c_m^2$ as required.
Using this bound in conjunction with the elementary series inequality $\sum_{j=1}^m j^{-1} \leq (1 + \log m)$, we have $K_m \geq 0$ and
\[
K_m =  \frac{1}{m}(c_m^2 - \Phi_m^2)\sum_{j=1}^{m}\frac{1}{j}\ \leq\   \frac{1}{m}c_m^2\sum_{j=1}^{m}\frac{1}{j}\ \leq \left(\frac{1+\log m}{m}\right)c_m^2
\]

An identical argument to that used between \eqref{eq: EC start} and \eqref{eq: EC end} shows that
$$
\mathbb{E}[C_{b,m,k}^2] = \frac{\log(n C_1)}{\gamma}
$$
and therefore 
$$
\mathbb{E}[c_m^2] \leq 2 C_{\mu,k}^2 + 2 \mathbb{E}[C_{b,m,k}^2] \leq 2 C_{\mu,k}^2 + \frac{2\log(bC_1)}{\gamma} .
$$
An identical argument to \eqref{eq: opt mmd start}-\eqref{eq: opt mmd end} gives that
$$
\mathbb{E}[\Phi_m^2] \leq \frac{\log(C_1)}{b \gamma} 
$$

From this the theorem follows by noting
\begin{align*}
 \mathbb{E}\left[ \mathrm{MMD}_{\mu,k}\left(\frac{1}{ms} \sum_{i=1}^m \sum_{j=1}^s \delta(x_{\pi(i,j)}^i) \right)^2 \right] = \frac{\mathbb{E}[a_m]}{(sm)^2} & \leq \mathbb{E}[\Phi_m^2] + \left(\frac{1+\log m}{m}\right) \mathbb{E}[c_m^2] \\
 & \leq \frac{\log(C_1)}{b \gamma} + 2 \left( C_{\mu,k}^2 + \frac{\log(bC_1)}{\gamma} \right) \left(\frac{1+\log m}{m}\right) . 
\end{align*}

\hfill $\square$

This argument relied on independence between mini-batches and therefore it may not easily generalise to the MCMC context.

\paragraph{Remarks:}
We observe that, in the myopic case only ($s=1$), one can alternatively recover Theorem \ref{thm: mini batch} as a consequence of Theorem 6 in \cite{chen2019stein}, once again using the observation that the kernel in \eqref{eq: k0 kernel} satisfies the preconditions of Theorem 6 in \cite{chen2019stein}.

The argument used to prove Theorem \ref{thm: mini batch} relies on independence between mini-batches and therefore it may not easily generalise to the MCMC context, in which this is unlikely to be true. Theorem 7 in \cite{chen2019stein} considered a particular form of dependence between mini-batches (once again, only for the case $s=1$), but this result does not directly apply to mini-batches sampled from MCMC output.

\subsection{Proof of \Cref{thm: sampled 2}} \label{app: proof of sampled 2}

The argument below is almost identical to that used in Theorem 2 of \cite{riabiz2020optimal}, with most of the effort required to handle the non-myopic optimisation having already been performed in \Cref{thm: nonmyopic fixed}.
In particular, it relies on the following technical result:

\begin{lemma}[Lemma 3 in \cite{riabiz2020optimal}]
			\label{Lemma-kernel-bounds}
			Let $\mathcal{X}$ be a measurable space and let $\mu$ be a probability distribution on $\mathcal{X}$.
			Let $k : \mathcal{X} \times \mathcal{X} \rightarrow \mathbb{R}$ be a reproducing kernel with $\int k(x,\cdot) \mathrm{d}\mu(x) = 0$ for all $x \in \mathcal{X}$.
			Consider a $\mu$-invariant, time-homogeneous reversible Markov chain $(x_i)_{i \in \mathbb{N}} \subset \mathcal{X}$ generated using a $V$-uniformly ergodic transition kernel, such that $V(x) \geq \sqrt{k(x,x)}$ for all $x \in \mathcal{X}$, with parameters $R \in [0,\infty)$ and $\rho \in (0,1)$ as in \eqref{eq: V unif ergod}. 
			Then we have that
			$$
			\sum_{i=1}^{n} \sum_{r \in \{1,\dots,n\} \setminus \{i\}}\mathbb{E} \left [k(x_i,x_r) \right] 
			\; \leq \;
			C_3 \sum_{i = 1}^{n - 1} \mathbb{E} \left[ \sqrt{k(x_i,x_i)} V(X_i) \right]
			$$
			with $C_3 := \frac{2 R \rho}{1 - \rho}$. \hfill $\square$
		\end{lemma}

The proof starts in a similar manner to the proof of \Cref{thm: sampled 1}, taking expectations of the bound obtained in \Cref{thm: nonmyopic fixed} to arrive at \eqref{eq: expect bound}.

An identical argument to that used in the proof of \Cref{thm: sampled 1} allows us to bound 
\begin{align*}
\mathbb{E}[C^2] \leq 2\left( C_{\mu,k}^2 + \frac{\log(n C_1)}{\gamma} \right) .
\end{align*}

Thus it remains to bound the first term in \eqref{eq: expect bound} under the assumptions that we have made on the Markov chain $(x_i)_{i \in \mathbb{N}}$.
To this end, we have that
\begin{align}
\mathbb{E}\left[ \underset{\substack{1^T\!w = 1 \\w_i\geq 0}}{\mathrm{min}}\ \mathrm{MMD}_{\mu,k} \left( \sum_{i=1}^n  w_i\delta(x_i) \right)^2 \right] & \leq \mathbb{E}\left[  \mathrm{MMD}_{\mu,k} \left( \frac{1}{n} \sum_{i=1}^n  \delta(x_i) \right)^2 \right] \nonumber  \\
& = \mathbb{E}\left[ \frac{1}{n^2} \sum_{i=1}^n\sum_{j=1}^n k(x_i,x_j) - \frac{2}{n}\sum_{i=1}^n\int k(x,x_i)\d \mu(x) + \iint k(x,y) \d \mu(x) \d \mu(y) \right] \nonumber \\
& = \mathbb{E}\left[ \frac{1}{n^2} \sum_{i=1}^n\sum_{j=1}^n k(x_i,x_j)  \right] \qquad \text{(since $\textstyle \int k(x,\cdot) \mathrm{d}\mu(x) = 0$)} \nonumber \\
& = \mathbb{E}\left[ \frac{1}{n^2} \sum_{i=1}^n k(x_i,x_i) \right] + \mathbb{E} \left[ \frac{1}{n^2} \sum_{i=1}^n \sum_{j \neq i} k(x_i,x_j)  \right] . \label{eq: MCMC bound}
\end{align}
The first term in \eqref{eq: MCMC bound} is handled as follows:
\begin{align*}
\frac{1}{n^2} \sum_{i=1}^n \mathbb{E}\left[ k(x_i, x_i) \right] & = \frac{1}{n^2} \sum_{i=1}^n \mathbb{E}\left[ \frac{1}{\gamma} \log e^{ \gamma k(x_i, x_i) } \right]  \\
				& \leq \frac{1}{\gamma n^2} \sum_{i=1}^n \log \left( \mathbb{E}\left[ e^{\gamma k(x_i, x_i) } \right] \right) \; \leq \; \frac{\log(C_1)}{\gamma n} 
\end{align*}
The second term in \eqref{eq: MCMC bound} can be controlled using \Cref{Lemma-kernel-bounds}: 
\begin{align*}
\mathbb{E} \left[ \frac{1}{n^2} \sum_{i=1}^n \sum_{j \neq i} k(x_i,x_j)  \right] \leq \frac{C}{n^2} \sum_{i = 1}^{n - 1} \mathbb{E} \left[ \sqrt{k(x_i,x_i)} V(X_i) \right] 
\leq \frac{C_3}{n^2} (n-1) C_2 \leq \frac{C_2 C_3}{n} .
\end{align*}

Thus we arrive at the overall bound
\begin{align*} 
\mathbb{E}\left[ \mathrm{MMD}_{\mu,k}\left(\frac{1}{ms} \sum_{i=1}^m \sum_{j=1}^s \delta(x_{\pi(i,j)}) \right)^2 \right] 
	&  \leq \frac{\log(C_1)}{n \gamma} + \frac{C_2 C_3}{n}  + 2 \left( C_{\mu,k}^2 + \frac{\log(n C_1)}{\gamma} \right) \left(\frac{1+\log m}{m}\right) ,
\end{align*}
as claimed. \hfill $\square$

\section{Semidefinite Relaxation} \label{appendix: sdr}

In this supplement we briefly explain how to construct a relaxation of the discrete optimisation problem \eqref{eq: optimal MMD}.
The standard technique for relaxation of a quadratic programme of this form is to construct an approximating semidefinite programme (SDP). 
This not only convexifies the problem but also replaces a quadratic problem in $v$ with a linear problem in a semidefinite matrix $M$.
To simplify the presentation we consider\footnote{The more general IQP setting, in which candidate points can be repeatedly selected, can similarly be cast as an SDP by proceeding with $s$ copies of the candidate set and $v \in \{0,1\}^{ns}$.} the BQP setting of \Cref{rem: binary}, so that $v \in \{0,1\}^n$. We also employ a change of variable $\tilde{v}_j := 2v_j-1$, so that $\tilde{v} \in \{-1,1\}^n$. 
By analogy with \eqref{eq: bqp formulation} we recast an optimal subset $\pi$ as the solution to the following BQP.
\begin{equation}
 \underset{\tilde v \in \{-1,1\}^n}{\mathrm{argmin}} \ \tilde v^\top K \tilde v + 2(\mathbf{1}^\top K+c_j^{i\top})\tilde v,\ \mathrm{s.t.\ }\mathbf{1}^\top \tilde v = 2s-n \label{eq: -1,1 formulation}.
\end{equation}
The relaxation treats $\tilde v$ as a continuous variable whose feasible set is the entire convex hull of $\{-1,1\}^n$. Define $\tilde V = \tilde v\tilde v^\top$ and then relax this non-convex equality, so that $\tilde V - \tilde v\tilde v^\top \succeq 0$ rather than the $\tilde V - \tilde v \tilde v^\top = 0$. Then rewrite this as a Schur complement, using the relation:
\[ M := \left(\begin{array}{ccc} 1 & \tilde v^\top \\ \tilde v & \tilde V \end{array}\right) \succeq 0 \iff \tilde V - \tilde v \tilde v^\top \succeq 0
\]
Consider now the two $(n+1) \times (n+1)$ matrices constructed as follows 
\[
A\! =\! \left(\begin{array}{cc}\!\!\! \mathbf{1}^\top K\, \mathbf{1} + 2c_j^{i\top}\!\!\!& \mathbf{1}^\top K + c_j^{i\top}\\ K\,\mathbf{1} +c_j^{i}   & K \end{array}\!\!\!\right) \quad 
B\!=\! \left(\begin{array}{ccc}0 & \frac{1}{2}\mathbf{1}^\top\!\!\! \\\!\!\! \frac{1}{2}\mathbf{1} &  \mathbf{0}\mathbf{0}^\top\!\!\! \end{array}\right)
\]
The SDP relaxation of \eqref{eq: -1,1 formulation} is then 
\begin{equation}\begin{aligned}
\mathrm{minimise}\ M \bullet A \quad \mathrm{s.t.}\ \ & \mathrm{diag}(M) = \mathbf{1} \\
& B \bullet M = 2s - n \\
& M \succeq 0 \\
\end{aligned} \label{eq:SDP relaxation in M} \end{equation}
($ X \bullet Y \equiv \sum\sum_{i,j=1}^n X_{ij}Y_{ij}$). Note that \eqref{eq:SDP relaxation in M} collapses to \eqref{eq: -1,1 formulation} when $\tilde{V} = \tilde{v} \tilde{v}^\top$ and $\tilde{v} \in \{-1,1\}^n$ are enforced.
Note that if the cardinality constraint $ B \bullet M = 2s - n$ is omitted, then \eqref{eq:SDP relaxation in M} is equivalent to the classical graph partitioning problem \emph{MAX-CUT} \citep{goemans1995improved}.

The SDP \eqref{eq:SDP relaxation in M} is linear in $M$ and is soluble to within any $\varepsilon > 0$ of the true optimum in polynomial time.
Its solution $M^\ast$, however, only solves the BQP \eqref{eq: -1,1 formulation} if  $\tilde  V^\ast = \tilde v^\ast \tilde v^{\ast\top}$, or equivalently $\mathrm{rank}(M^\ast) = 1$. 
This will not be true in general and the second part of a relaxation procedure is to round the output $\tilde v^\ast \in [-1,1]^n $ to a feasible vector $\tilde v \in \{-1,1\}^n$ for the BQP. 
\cite{goemans1995improved} introduced a popular randomised rounding approach for \emph{MAX-CUT}, and for the following exploratory simulations we adopted a similar approach. 
This starts by performing an incomplete Cholesky decomposition $\tilde V^\ast = UU^\top$ with $\mathrm{rank}(U) = r$. 
Since $\mathrm{diag}(\tilde V^\ast) =  \mathrm{1}$, the columns of $U$ all lie on the unit $r$-sphere.

To select exactly $m$ points we draw a random hyperplane through the origin of this sphere and translate it affinely until exactly $m$ points are separated from the rest (it is this translation that is a modification of the original approach for non-cardinality constrained problems, and which means the analysis of \cite{goemans1995improved} is not directly applicable). 
The resulting approximations are presented only as an empirical benchmark for Algorithms \ref{alg: myopic mmd}-\ref{alg: non-myopic mmd minibatch} and the detailed analysis of rounding procedures is well beyond the scope of this work.

We also find improved output by drawing $R>1$ points on the $r$-sphere and choosing the one for which the points separated off are best, in the sense of lowest cumulative KSD. This process imposes trivial additional computational cost. The semi-definite optimisations are performed using the \texttt{Python} optimisation package \texttt{MOSEK}.

\begin{figure*}[ht]
\centering
	\includegraphics[width=0.35\textwidth]{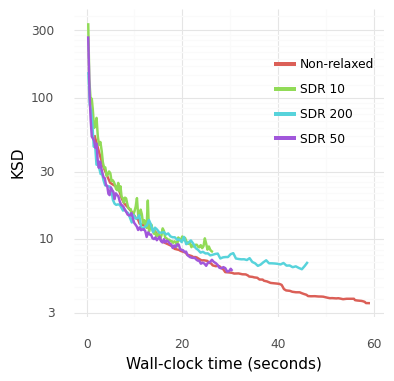}
	\includegraphics[width=0.35\textwidth]{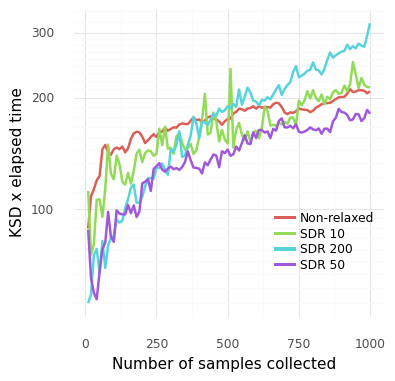}
	\caption{KSD  vs. wall-clock time, and time-adjusted KSD vs. number of selected samples, for the 4-dim Lotka--Volterra model also used in Section \ref{sec: empirical}, and with the same kernel specification. We draw 1000 samples using batch-size $b=100$ and choosing $s=10$ points simultaneously at each iteration. The four lines refer to the non-relaxed method (generated using the same code as in Figure \ref{fig: thinning}), as well as the approach employing semi-definite relaxation (taking the best of 10, 50 and 200 point selections, determined by drawing that many points on the sphere).} \label{fig: sdr}
\end{figure*}

Figure \ref{fig: sdr} shows that the semi-definite relaxation approach can be competitive in time-adjusted KSD. Each line in left pane represents the drawing of 1000 samples. The non-relaxed and best-of-50 SDR approaches closely mirror each other in time-adjusted KSD, though the non-relaxed approach is more efficient in that it achieves the same KSD in the same time with fewer samples chosen. Choosing $R>1$ imposes little additional computation time, leading to a performance improvement for $R=50$ over $R=10$, though past a certain point (visible here for $R=200$) this additional computation does become significant and harms performance.

\section{Choice of Kernel}  \label{sec: extra experiments}

As with all kernel-based methods, the specification of the kernel itself is of key importance. For the MMD experiments in Section \ref{subsec: BC}, we employed the squared-exponential kernel $k(x,y;\ell) = \exp(-\frac{1}{2}\ell^{-2}\Vert x-y\Vert^2 )$, and for the KSD experiments in Section \ref{subsec: MCMC} we followed \cite{chen2018stein,chen2019stein} and  \cite{riabiz2020optimal} and used the inverse multi-quadric kernel $k(x,y;\ell) = (1+\ell^{-2}\Vert x-y\Vert^2)^{-1/2}$ as the `base kernel' $k$ in \eqref{eq: oates unit ball} from  which the compound Stein kernel $k_\mu$ is built up.
The latter choice ensures that, under suitable conditions on $\mu$, KSD controls weak convergence to $\mu$ in $\mathcal{P}(\mathbb{R}^d)$, meaning that if $\text{MMD}_{\mu,k_\mu}(\nu) \rightarrow 0$ then $\nu \Rightarrow \mu$ \citep[Thm. 8]{gorham2017measuring}.

The next consideration is the length scale $\ell$. There are several possible approaches. For the simulations in Sections \ref{subsec: BC} and \ref{subsec: MCMC}, we use the median heuristic \citep{garreau2017large}. The length-scale $\hat{\ell}$ is calculated from the dataset themselves, using the formula $
\hat{\ell} = \sqrt{\tfrac{1}{2} \mathrm{Med}\{\Vert x_i - x_j \Vert^2 \}} $.
The indices $i,j$ can run over the entire dataset, or more commonly in practice, a uniformly-sampled subset of it. For the large datasets in Section \ref{sec: empirical}, we use 1000 points to calculate $\hat\ell$.

\begin{figure*}[ht]

	\includegraphics[width=0.28\textwidth]{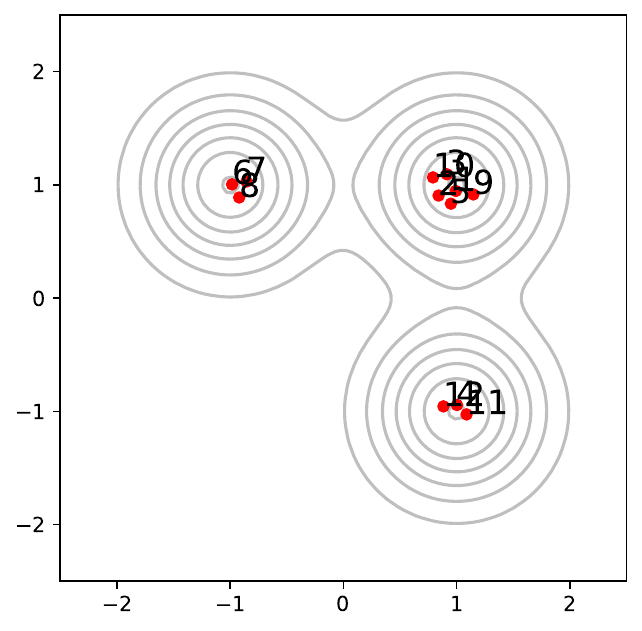}
	\includegraphics[width=0.28\textwidth]{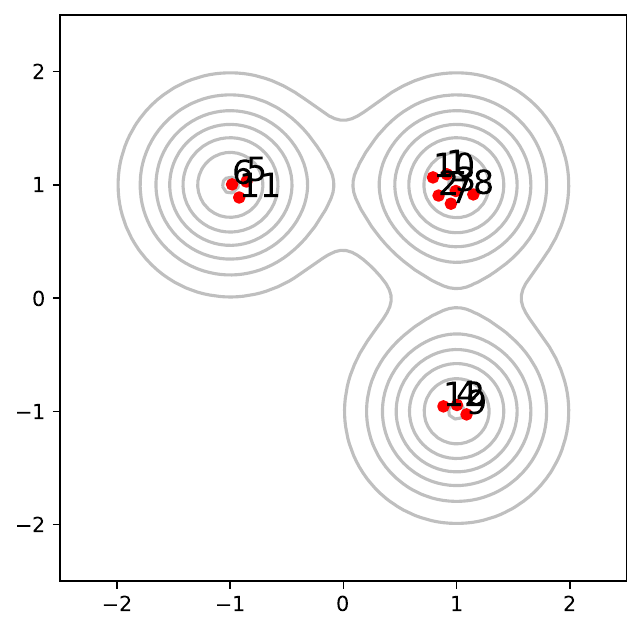}
	\includegraphics[width=0.28\textwidth]{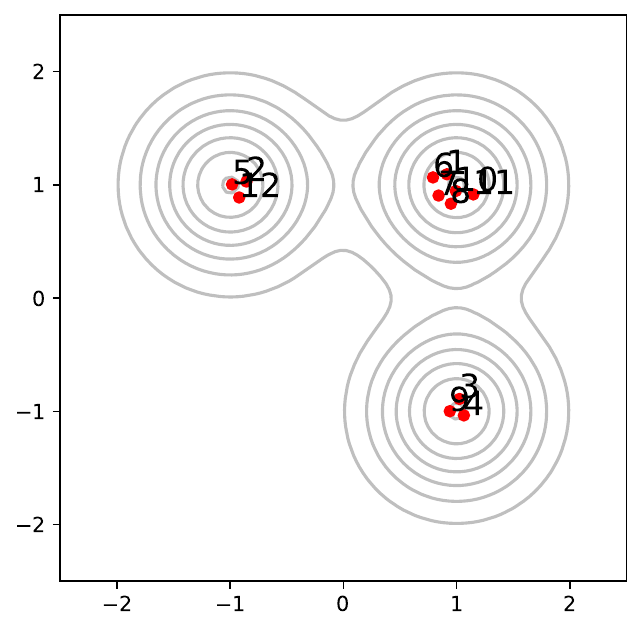}
	\includegraphics[width=0.28\textwidth]{images/fourplots_2}
	\includegraphics[width=0.28\textwidth]{images/fourplots_3}
	\includegraphics[width=0.28\textwidth]{images/fourplots_4}
	\includegraphics[width=0.28\textwidth]{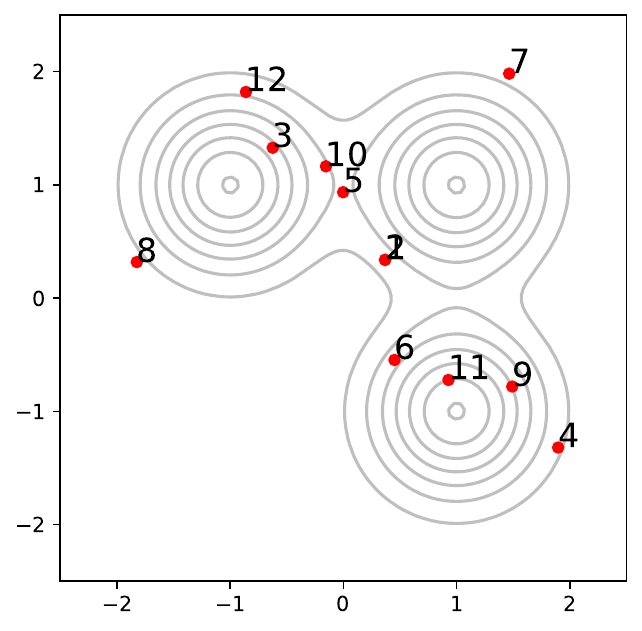}
	\includegraphics[width=0.28\textwidth]{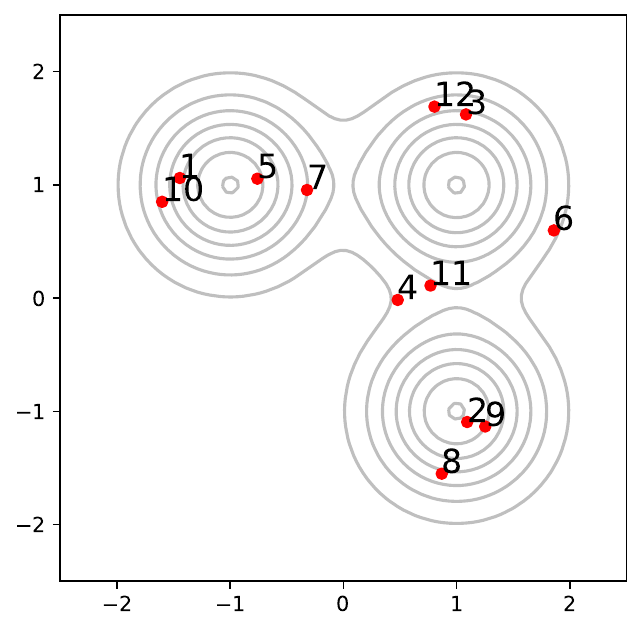}
	\includegraphics[width=0.28\textwidth]{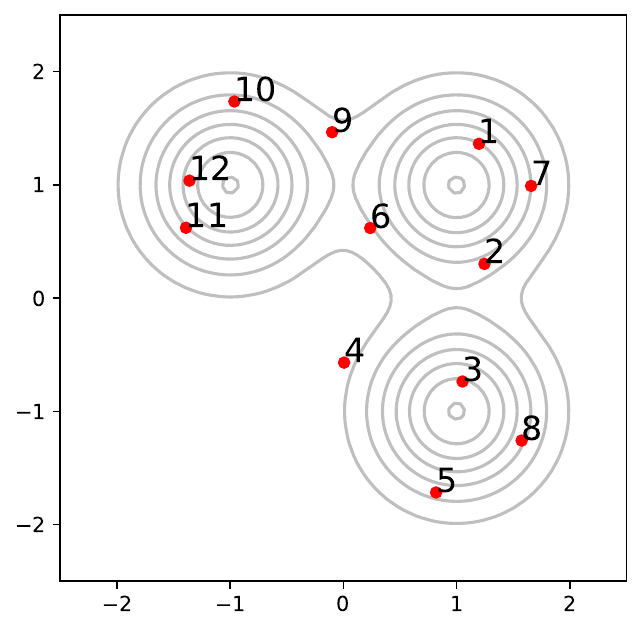}
	\centering
	\caption{Investigating the role of the length-scale parameter $\ell$ in the squared-exponential kernel $k(x,y;\ell) = \exp(-\frac{1}{2}\ell^{-2}\Vert x-y\Vert^2 )$. Model and simulation set-up as in Figure \ref{fig: visual example}. 
	Here 12 representative points were selected using the myopic method (left column), a non-myopic method (centre column), and by simultaneous selection of all 12 points (right column). 
	The kernel length-scale parameter $\tilde \ell$ set to $0.025$ (top row), $0.25$ (middle row; as Figure \ref{fig: visual example}) and $2.5$ (bottom row).
} \label{fig: kernel comparison}
\end{figure*}

To explore the impact of the choice of length scale on the approximations that our methods produce, in Figure \ref{fig: kernel comparison} we start with $\tilde\ell = 0.25$ (the value used to produce Figure \ref{fig: visual example} in the main text) and now vary this parameter, considering $0.1 \tilde\ell$ and $10\tilde\ell$. The difference in the quality of the approximation of $\nu$ to $\mu$ is immediately visually evident, even for such a simple model.
It appears that, at least in this instance, the median heuristic is helpful in avoiding pathologies that can occur when an inappropriate length-scale is used.

\vspace{2em}

\end{document}